\documentclass{article}
\usepackage{ijcai19}

\usepackage{times}
\usepackage{soul}
\usepackage{url}
\usepackage[hidelinks,draft]{hyperref}
\usepackage[utf8]{inputenc}
\usepackage[small]{caption}
\usepackage{graphicx}
\usepackage{amsmath}
\usepackage{booktabs}
\usepackage{algorithm}
\usepackage{xspace}
\usepackage{amssymb}
\usepackage{amsthm}
\usepackage{booktabs}
\usepackage{xcolor}
\usepackage{subfig}
\usepackage{multirow}

\newtheorem{theorem}{Theorem}
\newtheorem{lemma}[]{Lemma}

\DeclareMathOperator*{\argmax}{arg\,max}

\usepackage[noend]{algpseudocode}
\newcommand{\mb}{mini-batch\xspace}
\newcommand{\Mb}{Mini-batch\xspace}

\title{Submodular Batch Selection for Training Deep Neural Networks}

\author{
}

\author{
K J Joseph \and
Vamshi Teja R \and
Krishnakant Singh \and
Vineeth N Balasubramanian
\affiliations
Indian Institute of Technology Hyderabad
\emails 
\small \texttt{\{cs17m18p100001,ee15btech11023,cs15mtech11007,vineethnb\}@iith.ac.in
}}
\begin{document}

\maketitle

\begin{abstract}
Mini-batch gradient descent based methods are the de facto algorithms for training neural network architectures today.
We introduce a mini-batch selection strategy based on submodular function maximization. Our novel submodular formulation captures the informativeness of each sample and diversity of the whole subset. We design an efficient, greedy algorithm which can give high-quality solutions to this NP-hard combinatorial optimization problem. Our extensive experiments on standard datasets show that the deep models trained using the proposed batch selection strategy provide better generalization than Stochastic Gradient Descent as well as a popular baseline sampling strategy across different learning rates, batch sizes, and distance metrics.

%   Mini-batch gradient descent based methods are the de facto algorithms for training Neural Network architectures. Mini-Batch SGD algorithm are a  generalization of the full batch gradient descent where batch size is equal to the dataset size. The stochastic construction of \mb generates a high variance between the full batch gradient and \mb gradient. Leading to a much slower and poor convergence of the model.
%  In this work we formulate the problem of mini-batch selection as a constraint submodular maximization problem. We formulate a novel submodular objective which not only considers the \textit{informativeness} of each sample individually, but also \textit{diversity} of the mini-batch as a whole. We design an efficient, greedy algorithm which can give near optimal solution tot this NP-hard combinatorial optimization problem. Choice of the objective function is motivated by the need of fast creation of mini-batches for a fast training times. 

%   We compare our model with the 2 strong baselines, the models trained using the proposed result achieve a better generalization error in lesser number of epochs. Ablation studies are done to investigate the effects of the hyperparameters, proposed method is robust to the choice of hyperparameters in all the cases.   
\end{abstract}

\vspace{-10pt}
\section{Introduction} \label{sec:intro}

Deep learning methods are currently the state-of-the-art machine learning models for many applications, including computer vision, language understanding, and speech processing. The standard method for training deep neural networks is \mb stochastic gradient descent (SGD) while using backpropagation to compute the gradients.
% The \mb SGD is an optimization algorithm where mini-batches of data $D = \{D_1, \cdots D_t, \cdots, D_\mathcal{T}\}$ arrive one after the other. Here, each $D_t = \{d_1, d_2, \cdots, d_m\}$ is a \mb at time step $t$, containing $m$ examples sampled randomly from the dataset. 
The \mb SGD is an optimization algorithm where mini-batches of data $D_t = \{d_1, d_2, \cdots, d_m\}$ containing $m$ examples are sampled uniformly from the dataset $D$, at time $t$.
A loss function value w.r.t. the current model parameters $w_t$ is computed as $\mathcal{L}(w_t) = \sum_{i = 1}^{m}l(d_i|w_t)$ (where $l(.)$ is any differentiable loss function for the neural network), and the weights are updated to minimize $\mathcal{L}(w_t)$, according to the following equation:
% The \mb SGD is an optimization algorithm where mini-batches of $m$ data points $D_t$, is sampled uniformly from the dataset at time $t$.
% Then, its loss with respect to the current model parameters $w_t$ is calculated $\mathcal{L}(w_t) = \sum_{i = 1}^{m}l(d_i|w_t)$ and the weights are updated to minimize $\mathcal{L}(w_t)$, according to the following equation:
\begin{equation}
    \vspace{-3pt}
    w_{t+1} = w_t - \mu_t \frac{\partial \mathcal{L}(w_t)}{\partial w_t} 
    \vspace{-3pt}
\end{equation}
\noindent where $\mu_t$ is the learning rate at the $t^{th}$ step.

In this work, we hypothesize and validate that not only is the update of $w_t$ given $\frac{\partial \mathcal{L}(w_t)}{\partial w_t}$ crucial, but also the selection of the \mb $D_t$ used to compute the gradient. %$\frac{\partial \mathcal{L}(w_t)}{\partial w_t}$.
We formulate batch selection as solving a submodular optimization problem, which contributes to significant improvement in the generalization performance of the model. To the best of our knowledge, this is the first such effort on submodular importance sampling for SGD.

Each \mb selection is posed as a cardinality-constrained monotone submodular function maximization problem. %(explained further in Section \ref{sec:submodularity}).
This helps us leverage a greedy algorithm to solve this NP-hard combinatorial optimization problem, which guarantees a solution for a submodular objective function which is at least (in the worst case) $(1 - \frac{1}{e})$ (approximately 0.63) of the optimal solution \cite{nemhauser1978analysis}. %(Note that this is worst case behavior; in practice, the solution obtained is closer to the optimal.)

The key contribution of our work is a new submodular sampling strategy for \mb SGD, which helps train deep neural networks to have better generalization capability.
To achieve this, we formulate a submodular objective function, which takes into account the informativeness that each sample can add to the subset and at the same time ensure that the subset as a whole, is diverse. Further, we propose an efficient algorithm to scale to high sampling rates, as required for SGD while training neural networks.
% The key ingredient of such a submodular optimization problem is how the objective function $\mathcal{F}(.)$ is formulated. One of our key contribution in this work is to propose one such objective function which incorporates the characteristics of an ideal \mb into its formulation. 
% It takes into account the uncertainty of the model regarding a data point and strikes a balance between including a non-redundant data point and excluding an outlier. This naturally gives the model the expressive power to choose the apt set of data points to learn from, at any point in time. 
% We further contribute towards formulating an efficient subset selection algorithm that builds on the essence of  \cite{mirzasoleiman2013distributed} and \cite{mirzasoleiman2015lazier}, that scales the proposed approach to large datasets.
We conduct extensive experimental studies of the proposed submodular \mb selection methodology and show that it improves generalization capability of SGD as well as related previous efforts such as Loss based sampling \cite{loshchilov2015online}. We also show that the improved performance of the proposed methodology is consistent across different learning rates, mini-batch sizes and distance metrics. While submodular batch selection has been used extensively in the past for other settings such as active learning \cite{wei2015submodularity,chakraborty2015adaptive} and other methods such as Determinantal Point Process (DPP) \cite{zhang2017determinantal} have been used to diversify minibatch selection recently, to the best of our knowledge, this is the first submodular batch selection methodology\footnote{We use the terms batch selection and mini-batch selection interchangeably in this work.} for SGD while training neural networks. Importantly, our consistent increase in generalization performance over SGD is a notable achievement, which was missing in methods proposed earlier for mini-batch selection.

%the generalization performance of deep classification models when the proposed submodular \mb selection is used. Stochastic Gradient Descent (SGD) and Loss based sampling \cite{loshchilov2015online} are used as baselines to compare our proposed approach. When evaluated on three popular dataset; Street View House Numbers (SVHN) \cite{netzer2011reading}, CIFAR-10 and CIFAR-100 \cite{krizhevsky2009learning}, we consistently improve the classification accuracy on the unseen test images. We also note lower test loss on all datasets, which affirms better convergence and improved generalization capability while choosing the correct examples in each mini-batch. We perform ablation studies to show that the performance improvement of the proposed \mb selection scheme holds consistently across different learning rates, mini-batch sizes and different distance metrics.

% Concretely, the paper introduces the following contributions:

The remainder of this paper is organized as follows. We survey the literature related to our work in Section \ref{sec:related_work}; we review the concepts of submodularity, introduce our submodular function and describe our efficient implementation strategy in Section \ref{sec:submodular_selection}. The end-to-end algorithm is summarized in section \ref{sec:training}. The experimental setup, main results, and ablation studies are reported in Section \ref{sec:experiments}. We conclude with pointers for future work in Section \ref{sec:conclusion}.
\section{Related Work}\label{sec:related_work}
\vspace{-3pt}
Mini-batch selection strategies have been explored in convex settings in the past. \cite{zhao2014accelerating} proposed a sampling scheme based on partitioning data into balanced strata leading to faster convergence, while \cite{zhao2015stochastic} proved that the optimal sampling distribution is directly related to the absolute values of the gradient of the samples for convex objectives. However, the prohibitive cost of evaluating the gradient impedes their usage in practice. In non-convex settings, such as in deep neural networks, there have been fewer efforts for mini-batch selection, especially in the context of SGD. Extensions of \cite{zhao2015stochastic} to neural networks do not scale, due to the large number of trainable parameters in the deep models. Recently,  \cite{loshchilov2015online,alain2015variance} have tried to alleviate the cost of computing gradients by using loss-based sampling as an approximation to the gradients. Unfortunately, these methods are very sensitive to hyperparameters and perform inadequately in many cases \cite{katharopoulos2017biased}. The work was, in fact, validated only on MNIST data, which was acknowledged as a limitation of the work in \cite{loshchilov2015online}. The only other efforts to our knowledge consider more efficient approximations to the batch gradients i.e. variance of the samples \cite{chang2017active} or an upper bound on the gradient norm \cite{katharopoulos2018not}, but the generalization performance is only comparable to SGD in most cases. 
%shown to perform much better and are robust to hyper-parameters to a high degree. 
%Follow-up works such as \cite{chang2017active,katharopoulos2018not} show considerable improvements over \cite{loshchilov2015online}, but are only marginal better in terms of generalization error when compared with standard SGD, unlike our work. 
On the other hand, there have been efforts to speed up \mb SGD in general such as  \cite{allen2017katyusha,johnson2013accelerating}. However, these efforts do not focus on batch selection or importance sampling, and show lower performance than existing importance sampling methods as noted in \cite{katharopoulos2018not}.

Submodular optimization has been successfully applied to varied tasks like document summarization \cite{lin2011class}, sensor placement \cite{shamaiah2010greedy}, speech recognition systems \cite{wei2014submodular}, to name a few. However, there has been no work so far on using submodularity for batch selection. 
% making them a ideal choice for summarizing datasets, they have found comprehensive usage in NLP \cite{lin2011class}, video \cite{gygli2015video} and many other domains.
\cite{das2008algorithms} proved that variance between a predictor variable using full batch and a \mb is submodular. We were initially motivated by this observation to propose a submodular batch selection strategy for SGD.  
%These works leads us to hypothesise that the submodular optimization would be a perfect fit for \mb selection for SGD. 
The existing efforts that are closest to ours include  Determinantal Point Process (DPP) \cite{zhang2017determinantal}, Repulsive Point Processes (RPP) \cite{zhang2018active} and \cite{singh2018submodular}. Both DPP and RPP can be considered a special case of probabilistic  submodular functions (PSF), although not explicitly called so in their work. However, these methods are computationally  inefficient. Even  with faster versions, they are prohibitively costly to be applied in deep neural networks \cite{li2016fast}. \cite{singh2018submodular} attempt a similar objective, but the objective considered is truly not submodular, and their results are largely inconclusive.
%Our work attempts to deal with these deficiencies by creating a submodular function that can be evaluated much more efficiently and also provides a better performance than existing methods.
Another body of work that can be considered close to our efforts are those of self-paced learning \cite{thangarasa2018self} and curriculum learning \cite{zhou2018minimax}. However, their objectives are different, and one can consider using our batch selection strategy along with any such method too.

%These works are generally based on a student teacher network architecture \cite{thangarasa2018self}, or training a proxy model \cite{coleman2018select}. These methods are orthogonal to our work and can used in conjunction with our method, for example training the teacher or proxy networks to obtain better results.

%To the best of our knowledge, we are the first to use submodular optimization to select mini-batches to train neural networks. The key challenge here is make it practical is to cope up with the very high sampling rate involved. The efficient algorithm, explained in Section \ref{sec:improving_efficiency}, makes our method viable.

% \paragraph{Sample selection for better training of neural networks}
% {...in progress...}

\section{Submodular Batch Selection Methodology} \label{sec:submodular_selection}
We begin this section with a brief introduction to submodularity, before presenting our methodology.
%In this section we first review the notion of submodular functions and show how a subset which maximizes such a function would imply it being diverse. We formulate the objective function and introduce the subset selection algorithm in the subsequent sub-sections.
% Next we formulate a submodular function that works best for \mb selection for gradient descent. We also discuss how we can scale up the submodular \mb selection to large datasets in Section \ref{sec:improving_efficiency}.
% and discuss some of its theoretical guarantees in Section \ref{sec:var_reduction}.

\subsection{Submodularity} \label{sec:submodularity}
% Submodularity is a well studied topic, extensively used in operations research, game theory, economics and combinatorial optimization. 
Given a finite set $V = \{1,2,\cdots,n\}$, a discrete set function $\mathcal{F}:2^V \to \mathbb{R}$, that returns a real value for any subset $S \subseteq V$ is \textit{submodular} if 
\begin{equation}
    \mathcal{F}(A) + \mathcal{F}(B) \geq \mathcal{F}(A \cup B) + \mathcal{F}(A \cap B) \quad \forall A,B \subseteq V
\end{equation}
\noindent A more intuitive way of defining a submodular function is in terms of the marginal gain of adding a new element to a subset. Let $\mathcal{F}(e|S) = \mathcal{F}(e \cup S) - \mathcal{F}(S)$ denote the marginal gain of adding an element $a$ to $S$. $\mathcal{F}$ is \textit{submodular} if 
\begin{equation}
\mathcal{F}(a|S) \geq \mathcal{F}(a|T) \qquad  \forall S \subseteq T \subseteq V\setminus a 
\end{equation}
\noindent This is also called the \textit{diminishing returns property}, where the incremental gain of adding a new element to a set decreases as the set grows from $S$ to $T$. Hence, a subset that maximizes a submodular objective function $\mathcal{F}(.)$ would have least redundant elements over other subsets of the same cardinality because any redundant element will reduce the value of the submodular objective function.

A function is \textit{monotone non-decreasing} if $\forall A \subseteq B, \mathcal{F}(A) \leq \mathcal{F}(B)$. $\mathcal{F}(.)$ is said to be \textit{normalized} if $\mathcal{F}(\emptyset) = 0$. A greedy algorithm \cite{nemhauser1978analysis} can be used to maximise a normalized monotone submodular function with cardinality constraints, with a worst-case approximation factor of $1 - \frac{1}{e}$. An instance of such a greedy algorithm can be as follows: In the $i^{th}$ iteration, the algorithm selects an item $s_i$ that maximizes the conditional gain, i.e. $s_i = \argmax_{a \in V \setminus S_{i-1}} \mathcal{F}(a | S_{i-1})$. The subset $S_i$, initially empty, is updated as: $S_i \gets \{s_i\} \cup S_{i-1}$. The algorithm terminates when $S_i$ meets the cardinality constraint $|S_i| \leq k$.

\subsection{Submodular Batch Selection} 
\label{sec:objective}
% The key component in the submodular optimization is the objective function. Here, we formulate one such function that takes into account the informativeness that each of the data-point can add to the subset and ensures that the subset as a whole is diverse. We prove that the objective is submodular and is monotonically non-decreasing, so that the greedy algorithm can be applied.

% Here, we formulate one such function that takes into account the informativeness that each of the data-point can add to the subset and ensures that the subset as a whole is diverse. We prove that the objective is submodular and is monotonically non-decreasing, so that the greedy algorithm can be applied.

% We formulate the objective function denoting the score for a mini-batch S, as follows:
% \begin{equation}
%     \mathcal{F}(S) = \sum_{x_i \in S} \lambda_1 U(x_i) + \lambda_2 R(x_i) + \lambda_3 MC(x_i) + \lambda_4 FM(x_i)
%     \label{eqn:full}
% \end{equation}

% A \mb $S$ that maximizes this score would be selected in each training iteration. Each term in the objective function is explained below:
An appropriate batch selection strategy for \mb SGD would need to consider multiple criteria to choose the most relevant samples. A primary criterion we consider is that each selected sample must be as informative as possible. We use the model uncertainty as the measure of informativeness.

%an ideal metric to measure this. Maximizing the following score would encourage informative samples to be included in the \mb:
%The key component in the submodular optimization is the objective function. In this section, we explain how we come up with an objective function that would model the characteristics of an ideal mini-batch. 
%should be as informative as possible. The uncertainty of the model would be an ideal metric to measure this. Maximizing the following score would encourage informative samples to be included in the \mb:
\vspace{2mm}
\hspace{-10pt}\textbf{Uncertainty Score} [ $U(x_i)$]: 
The uncertainty of each data point is computed as the entropy of the current model $w^t$ at training iteration $t$. $C$ is the set of all classes. This allows the model to select the samples that confuses it the most in a \mb:
\begin{equation}
    U(x_i) = - \sum_{y \in C} P(y|x_i, w^t) \log P(y|x_i, w^t)
     \label{eqn:Uncertainity_Score}
\end{equation}

A subset that maximizes only the Uncertainty Score, would potentially lead to the inclusion of similar data points with high entropy in the \mb. This redundancy should be avoided to make the \mb diverse. The following score helps to contain the inclusion of redundant data points:

\vspace{2mm}
\hspace{-10pt}\textbf{Redundancy Score} [ $R(x_i)$]: Two data points $x_i$ and $x_j$ may separately furnish valuable information, but including both may make the subset less maximally informative. We use Redundancy Score to take this into account. Given $\phi(.)$ to be any distance metric between the two data points, a greater value of the minimum distance between points in the subset would imply more diversity among the data points in the subset. (Needless to say, this score is dependent on the choice of distance metric, and we study this in our experiments.)
\begin{equation}
    R(x_i) = \min_{x_j \in S : i \neq j} \phi(x_i, x_j)
    \label{eqn:Redundancy_Score}
\end{equation}

Going further, one can notice that outlier samples may maximize the above scores. In order to counter such a selection of batches, we introduce the following score.
%A soiled way that to maximize an objective function composed of Uncertainty score and Redundancy score would be to include outliers in the \mb. This is alleviated in:

\vspace{2mm}
\hspace{-10pt}\textbf{Mean Closeness Score} [$MC(x_i)$]: %It keeps a check on outliers to be excluded in the \mb. 
This term encourages the selection of data points that are closer to the mean of all the examples ($\mu = \frac{1}{|V|}\sum_{k = 1}^{|V|} x_k$) to be picked. This avoids the selection of outlier samples to the extent possible.
% $\phi(.)$ is the distance metric employed.
\begin{equation}
    MC(x_i) = \phi(x_i, \mu)
     \label{eqn:Mean_Closeness_Score}
    \vspace{0pt}
\end{equation}

Finally, there has been recent work to show that closeness in the feature space of a deep neural network may be a better indicator of how similar two samples are (as shown by \cite{wei2014submodular} in the speech domain). We hence also include a term to explicitly enforce diversity in the feature space of the given data. 

\vspace{2mm}
\hspace{-10pt}\textbf{Feature Match Score} [$FM(x_i)$]: This score selects samples that are diverse across each dimension in the feature space. %This score function is motivated by feature-based submodular function \cite{wei2014submodular}, used in the speech domain. 
$g(.)$ is a non-negative monotone non-decreasing concave function, $U$ is a set of fixed features and $m_u(x_i)$ is a non-negative score, measuring the degree to which data point $x_i$, possesses the feature $u$. 
\begin{equation}
    FM(x_i) = \sum_{u \in U} g(m_u(x_i))
    \label{eqn:feature_match}
    \vspace{-5pt}
\end{equation}
Our implementation of this score (as well as others) is described in Section \ref{sec:r_implementation_details}.

\noindent We combine the abovementioned scores to form our objective function for batch selection, $\mathcal{F}(S)$, as below and then show the submodularity of the proposed $\mathcal{F}(S)$.
\begin{equation}
    \mathcal{F}(S) = \sum_{x_i \in S} \lambda_1 U(x_i) + \lambda_2 R(x_i) + \lambda_3 MC(x_i) + \lambda_4 FM(x_i)
    \label{eqn:full}
\end{equation}

% These four score functions are carefully selected so that maximizing $\mathcal{F}(S)$ (Equation \ref{eqn:full}) would lead to a rich \mb. The \textit{Redundancy Score} and \textit{Feature Match Score} ensures that the subset is diverse while \textit{Uncertainty Score} ensures that the most informative examples would be favoured into to the \mb. \textit{Mean Closeness Score} ensures that outliers wont be added in. Four trade-off parameters ($\lambda_1, \lambda_2, \lambda_3, \lambda_4$) control the relative importance of each of the term.
% Carefull ablation studies on the contribution of each of the parameters is carried out in Section \ref{sec:r_trade_off_param}.
% More implementation specific details of each of the score function is discussed in Section \ref{sec:r_implementation_details}.
\noindent Given a dataset with $N$ training data points, a \mb of size $k$ is selected by solving the following cardinality-constrained submodular optimization problem:
\begin{equation}
    %\vspace{-4pt}
    \max_{S \subseteq V, |S| \leq k} \mathcal{F}(S)
    \label{eqn:smdl_optimizer}
    \vspace{-5pt}
\end{equation}

\noindent We now show that the score function $\mathcal{F}$ is indeed submodular and is monotonically non-decreasing. This would allow us to solve the problem in (\ref{eqn:smdl_optimizer}) using a greedy approach \cite{nemhauser1978analysis}.

%The scale of the problem calls for much more efficient ways of solving Equation \ref{eqn:smdl_optimizer}, which we discuss in Section \ref{sec:improving_efficiency}. More implementation specific details of each of the score function is discussed in Section \ref{sec:r_implementation_details}. An ablation study on the contribution of each of the score function is given in Section \ref{sec:r_trade_off_param}.

\begin{lemma}
The score function $\mathcal{F}(.)$, defined in Eqn \ref{eqn:full} is submodular.
\end{lemma}
\vspace{-12pt}
\begin{proof}
Consider two subsets of training examples from a dataset $V=\{x_1, x_2, \cdots, x_N\}$; $S_1$ and $S_2$, such that $S_1 \subseteq S_2 \subseteq V$. Let $a$ be an element not selected so far: $a \in V \setminus S_2$. The marginal gain of adding $a$ to $S_1$ is given by:
\begin{equation*}
    \begin{split}
        \mathcal{F}(a|S_1) 
        =~ & \mathcal{F}(\{a\} \cup S_1) - \mathcal{F}(S_1) \\
        =~ & \lambda_1 U(a) + \lambda_2 \min_{a_j \in S_1} \phi(a, a_j) + \lambda_3 MD(a) + \lambda_4 FM(a)
    \end{split}
\end{equation*}
Similarly, the marginal gain of adding $a$ to $S_2$ is given by:
\begin{equation*}
    \begin{split}
        \mathcal{F}(a|S_2) 
        =~ & \mathcal{F}(\{a\} \cup S_2) - \mathcal{F}(S_2) \\
        =~ & \lambda_1 U(a) + \lambda_2 \min_{a_j \in S_2} \phi(a, a_j) + \lambda_3 MD(a) + \lambda_4 FM(a)
    \end{split}
\end{equation*}
Since $S_1 \subseteq S_2$, the minimum distance of the new point $a$, from $S_1$ would be greater than any element from $S_2$, as there may exist a point in $S_2$ that is much closer to $a$, than any element from its subset $S_1$. Hence, 
\begin{equation*}
    \min_{a_j \in S_1} \phi(a, a_j) \geq \min_{a_j \in S_2} \phi(a, a_j)
\end{equation*}
Thus, we can claim $\mathcal{F}(a|S_1) \geq \mathcal{F}(a|S_2)$. Hence, the score function $\mathcal{F}(.)$ is submodular.
\end{proof}

\begin{lemma}
The score function $\mathcal{F}(.)$ in Eqn \ref{eqn:full} is a monotonically non-decreasing function.
\end{lemma}
\vspace{-8pt}
\begin{proof}
Consider a subset $S$ and an element $a \in V \setminus S$. When $a$ is added to $S$, the function value of $\mathcal{F}(\{a\} \cup S)$ changes by $\lambda_1 U(a) + \lambda_2 \min_{a_j \in S} \phi(a, a_j) + \lambda_3 MD(a) + \lambda_4 FM(a)$. All these are non-negative quantities. Thus, $\mathcal{F}(\{a\} \cup S) \geq \mathcal{F}(S)$, and the score function $\mathcal{F}(.)$ is hence monotonically non-decreasing.
\end{proof}

\begin{theorem}
Let $S^{*}$ denote the optimal solution of the problem in (\ref{eqn:smdl_optimizer}) and $S$ denote the solution obtained for the same problem using a greedy approach. Then:
\begin{equation*}
    \mathcal{F}(S) \geq (1 - \frac{1}{e}) \mathcal{F}(S^{*})
    \vspace{-2pt}
\end{equation*}
\label{theorem1}
\end{theorem}
\vspace{-16pt}
\begin{proof}
Having proved that score function $\mathcal{F}(.)$ is submodular in Lemma $1$ and that it is monotonically non-decreasing in Lemma $2$, the proof follows directly from Theorem $4.3$ in \cite{nemhauser1978analysis}. 
\end{proof}

\subsection{Scaling to High Sampling Rates}\label{sec:improving_efficiency}
It is interesting to note that application settings where submodularity has worked well hitherto \cite{wei2014submodular,wei2015submodularity,chakraborty2015adaptive,lin2011class} do not require a high sampling rate, 
as much as demanded by mini-batch selection in SGD. Consider a mini-batch training algorithm that consists of $p$ epochs and $q$ iterations in each epoch, the submodular batch selection needs to be carried out $p \times q$ times. Concretely, for training on the CIFAR-100 dataset having 50,000 examples and a batch size of 50 ($q$ is hence 50000 / 50), we need 100,000 batch selections for 100 epochs ($p = 100$). Even a greedy algorithm \cite{nemhauser1978analysis} that uses scores such as pairwise distance metrics has a complexity of $O(n^2)$, where $n$ is the dataset size. This would be too slow for use with SGD. We hence seek more efficient mechanisms to implement the proposed batch selection strategy.
%Hence it would not scale to the requirements of \mb selection.

Recent efforts have attempted to make submodular sampling faster in a more general context (not in SGD), such as Lazy Greedy \cite{minoux1978accelerated}, Lazier than Lazy Greedy (LtLG) \cite{mirzasoleiman2015lazier}, and Distributed Submodular Maximization  \cite{mirzasoleiman2013distributed}.
In this work, we present a new methodology for efficient submodular sampling inspired by Distributed Submodular Maximization algorithm and Lazier than Lazy Greedy algorithm (LtLG), as described in Algorithm \ref{algo:Sampling Algorithm}. The algorithm partitions the training set $V$ into $m$ partitions in Line 2 and runs Lazier than Lazy Greedy algorithm (LtLG) \cite{mirzasoleiman2015lazier} (described later in this section) on each partition (Lines 5 through 8) to obtain $m$ subsets, each of size $b$. These subsets ($S_is$) are then merged in Line 9. The final subset $S$ is selected from this merged set by running LtLG again on it. This divide-and-conquer strategy is motivated by the Distributed Submodular Maximization algorithm  \cite{mirzasoleiman2013distributed}.
\begin{algorithm}
\caption{Algorithm \textsc{GetMiniBatch}}
\label{algo:Sampling Algorithm}
\begin{algorithmic}[1]
\Require{Training set $V$, Model at $k^{th}$ iteration $w_k$, Batch size $b$, Number of partitions $m$, $\mathcal{F}: 2^V \to \mathbb{R}$ (Eqn \ref{eqn:full}).}
\Ensure{\Mb $S \subseteq V$ satisfying $|S| \leq b$.}

\State $S \leftarrow \phi$ 
\State Partition $V$ into $m$ sets $V_1, V_2, V_3, \cdots, V_m$.
\For{i = 1 to m} 
\State $S_i \leftarrow \phi$
\For{j = 1 to b} \Comment \textit{Do LtLG for each partition.}
\State $R \leftarrow$ a subset of size $s$ obtained by sampling randomly from $V_i \setminus S_i$.
\State $a_j \leftarrow \argmax_{a \in R} \mathcal{F}(a|S_i)$
\State $S_i \leftarrow S_i \cup \{a_j\}$
\EndFor
\EndFor
\State $S_{merged} \leftarrow \bigcup_{i=1}^{m}S_i$ \Comment \textit{Merge result of each partition.}
\For{j = 1 to b} \Comment \textit{Do LtLG on the merged set.}
\State $R \leftarrow$ a subset of size $s$ obtained by sampling randomly from $S_{merged} \setminus S$.
\State $a_j \leftarrow \argmax_{a \in R} \mathcal{F}(a|S)$
\State $S \leftarrow S \cup \{a_j\}$
\EndFor
\State Return $S$
\end{algorithmic}
\end{algorithm}

The Lazier than Lazy Greedy (LtLG) \cite{mirzasoleiman2015lazier} algorithm starts with an empty set and adds an element from set $R$, which maximizes the marginal gain $\mathcal{F}(a|S)= \mathcal{F}(a \cup S) - \mathcal{F}(S)$. This is repeated until the cardinality constraint ($|S| \leq b$) is met. The set $R$ is created by randomly sampling $s= \frac{|V|}{b}log\frac{1}{\epsilon}$ items from the superset V, where $\epsilon$ is a user-defined tolerance level. We refer the readers to \cite{mirzasoleiman2015lazier} for further information. The model at the $k^{th}$ training iteration, $w_k$, is used while computing $\mathcal{F}(.)$ as defined in Equation \ref{eqn:full}.

The solution produced by Algorithm \ref{algo:Sampling Algorithm}, $S$, has the following approximation guarantee with the optimal solution $S^{*}$:
\begin{equation}
     \mathcal{F}(S) \geqslant \frac{(1-e^{-1})^2}{min(m, b)}(1-e^{-1}-\epsilon) \mathcal{F}(S^{*})
\end{equation}
Here, $m$ refers to the number of partitions, $b$ refers to the mini-batch size and $e$ is the base of natural logarithm. Our empirical results (Section \ref{sec:r_learning}) shows that the approximation is much better than this lower bound in practice.

% \subsection{Achieving Variance Reduction} \label{sec:var_reduction}

\subsection{Gradient Descent with Submodular Batches}
\label{sec:training}
The proposed batch selection strategy can work with any \mb gradient descent based optimization algorithms. Algorithm \ref{algo:Training Algorithm} summarizes the end-to-end training procedure.

\begin{algorithm}
\caption{Algorithm \textsc{Submodular SGD}}
\label{algo:Training Algorithm}
\begin{algorithmic}[1]
\Require{Training Set $V$, Optimizer $\pi(.,\eta)$, \# of epochs p, Batch size $b$, \# of partitions $m$.}
\Ensure{Trained model $w_p^{\frac{|V|}{b}}$.}

\State $\tau \leftarrow 1$
\State Initialize the model $w^1_\tau$.
\For{i = 1 to p}
% \State $V_{epoch} \leftarrow V$
\For{j = 1 to $\frac{|V|}{b}$}
\State $S \leftarrow \textsc{GetMiniBatch}(V, w^j_\tau, b, m)$
\State $\nabla J(w^j_\tau) \leftarrow \frac{\partial}{\partial w} \sum_{k \in S} L(y_k, f(x_k, w^j_\tau))$
\State $w^{j+1}_\tau \leftarrow w^j_\tau + \pi(\{w^{1:j}_\tau\}, \{\nabla J^{1:j}_\tau\}, \eta)$
% \State $V_{epoch} \leftarrow V_{epoch} \setminus S$
\EndFor
\State $w^1_{\tau + 1} \leftarrow w^{j+1}_\tau$; $\tau \leftarrow \tau + 1$
\EndFor
\State return $w^{\frac{|V|}{b}}_{p}$
\end{algorithmic}
\end{algorithm}

\noindent Within each iteration in each epoch, a \mb $S$ is selected using Algorithm \ref{algo:Sampling Algorithm} (Line 5). Lines 6 and 7 update $w$ with a gradient descent optimizer $\pi(.,\eta)$, consuming the current set $S$. $\eta$ is the learning rate. Any differentiable loss function can be used as $L(.)$. Momentum-based and adaptive learning rate-based gradient descent methods could be used to further improve the learning based on submodular batches.

%%%%%%%%% FIGURE STARTS
\begin{figure*}
\vspace{-40pt}
\centering
\subfloat[Test Error on SVHN]{\includegraphics[width=0.30\linewidth]{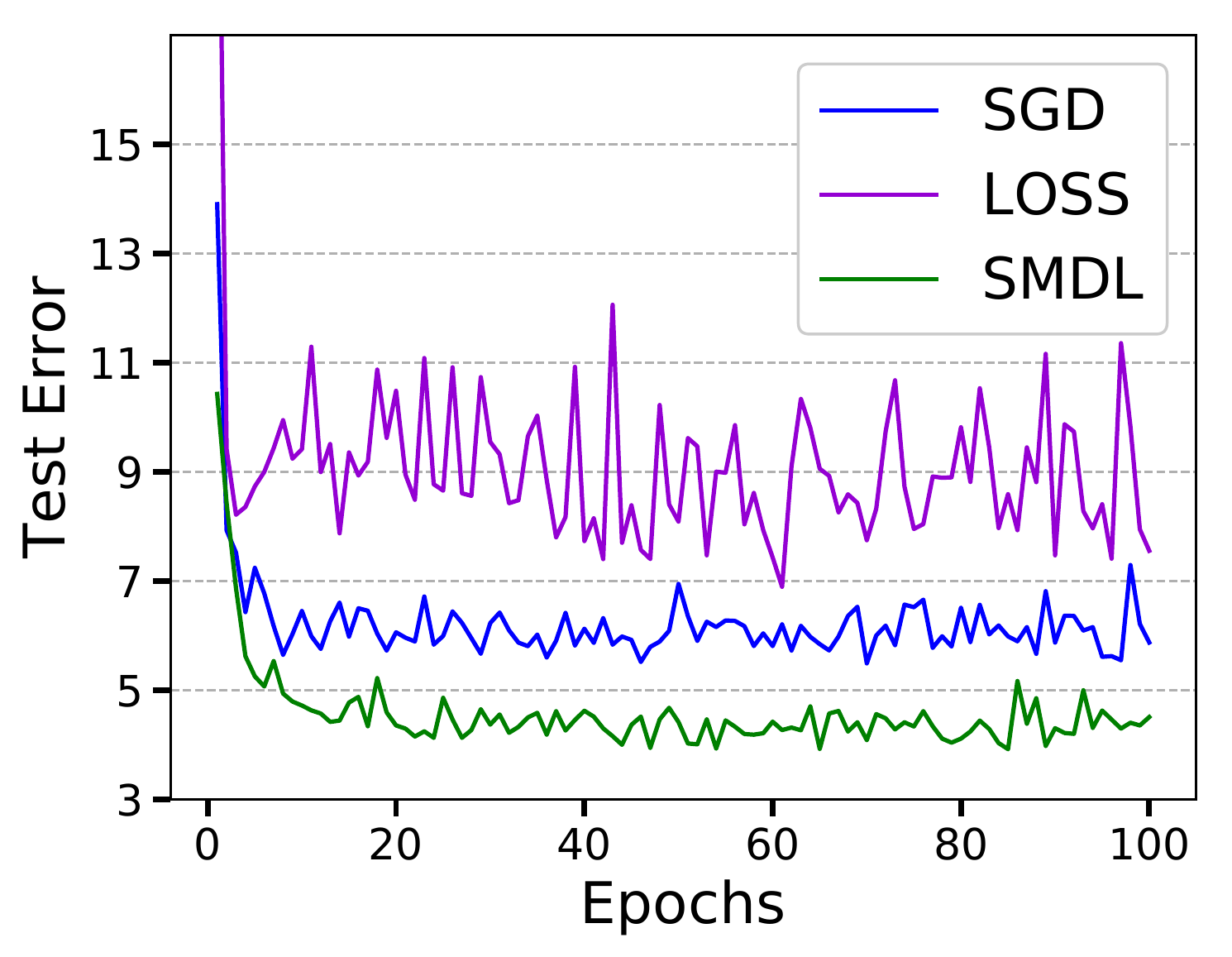}}
\hspace{15pt}\subfloat[Test Error on CIFAR-10]{\includegraphics[width=0.30\linewidth]{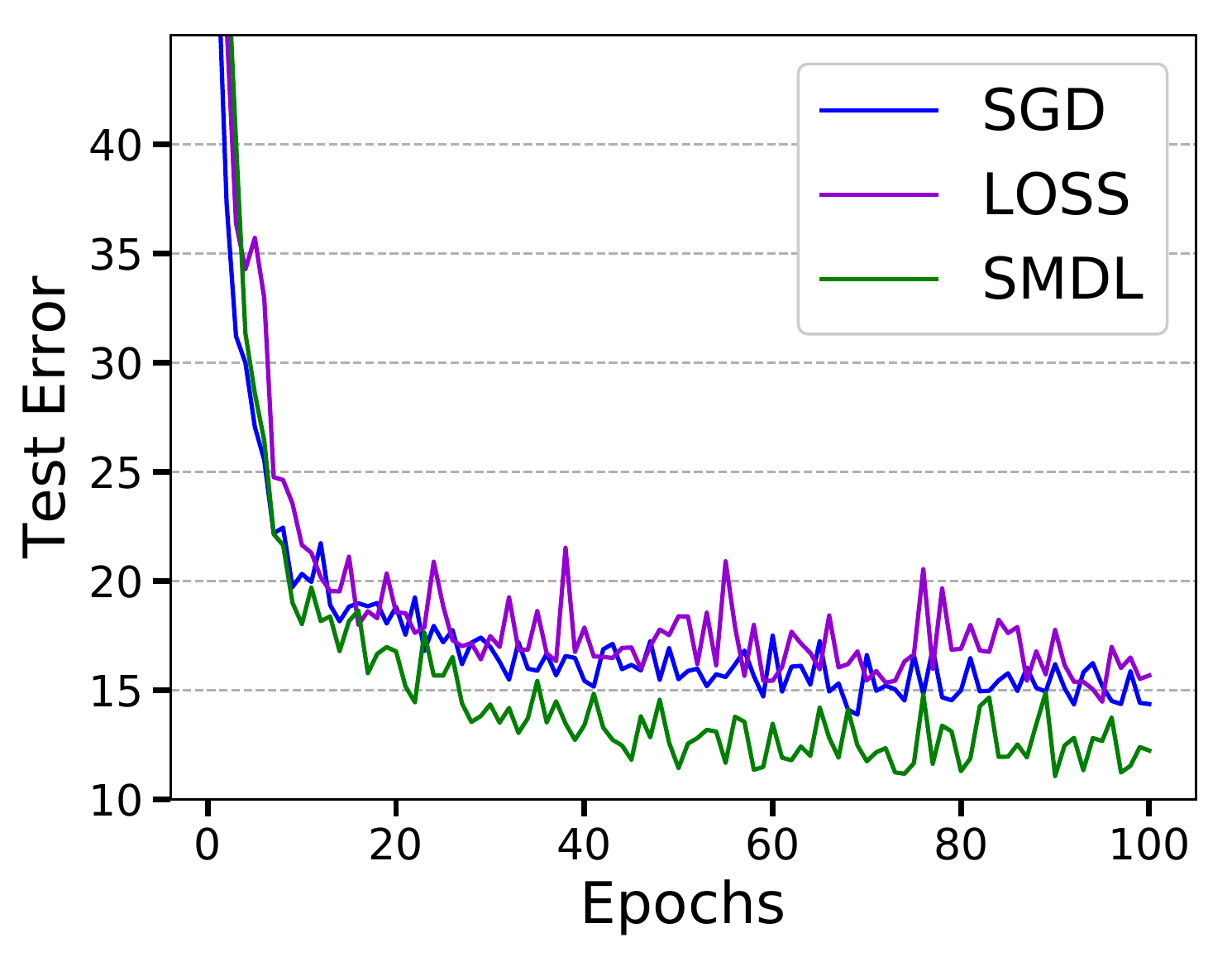}}
\hspace{15pt}\subfloat[Test Error on CIFAR-100]{\includegraphics[width=0.30\linewidth]{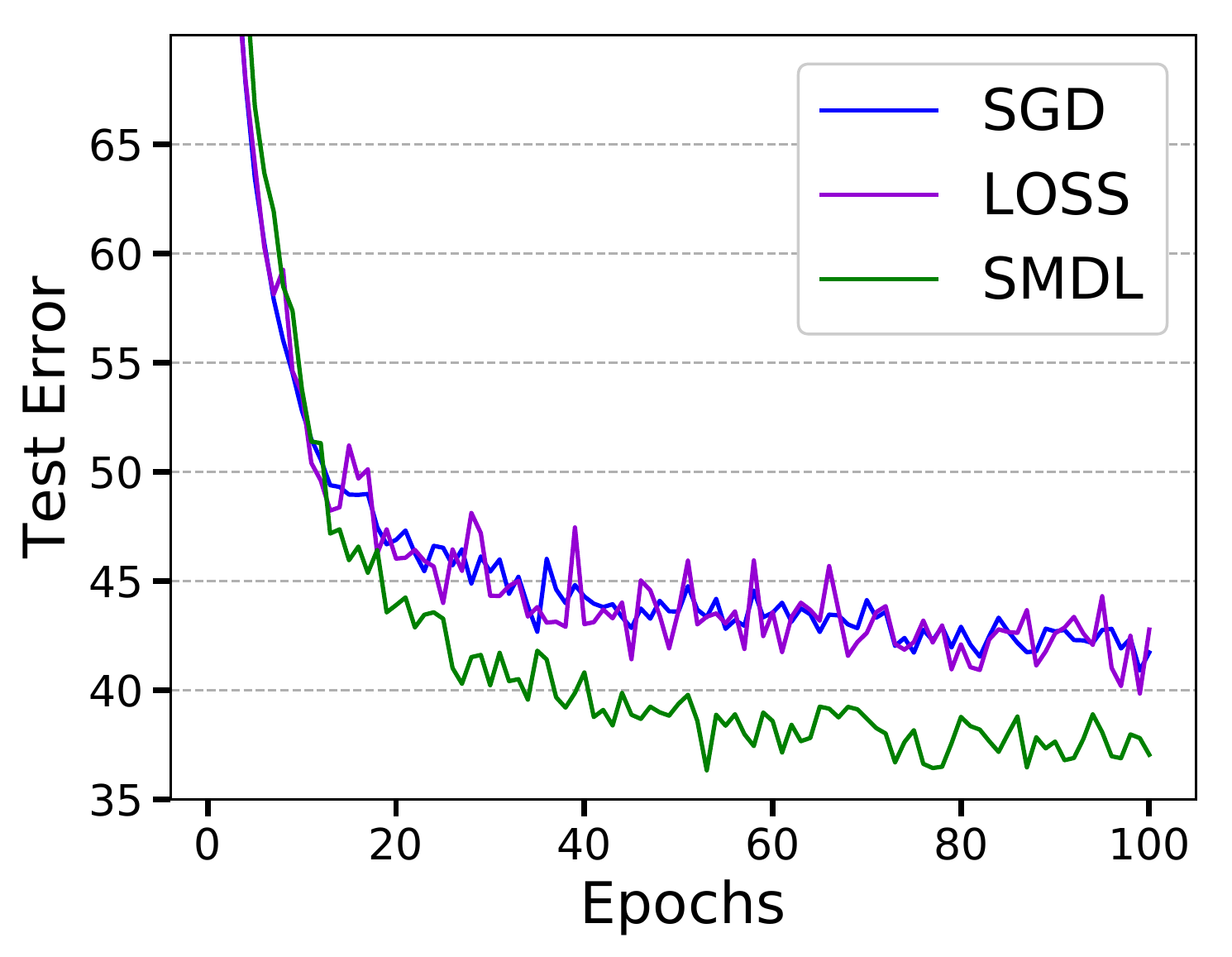}} 
\vspace{-9.6pt}
\subfloat[Test Loss on SVHN]{\includegraphics[width=0.30\linewidth]{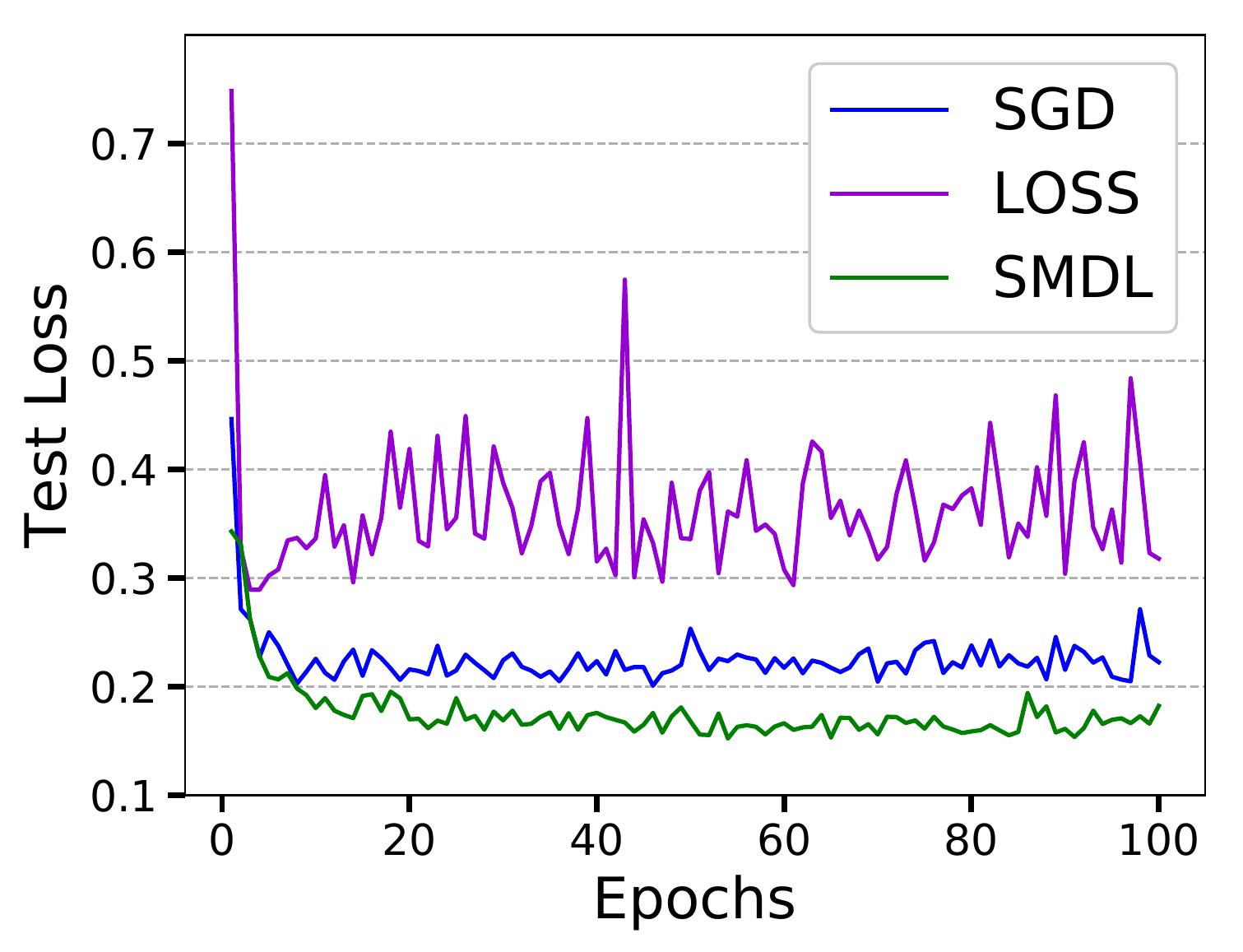}}
\hspace{15pt}\subfloat[Test Loss on CIFAR-10]{\includegraphics[width=0.30\linewidth]{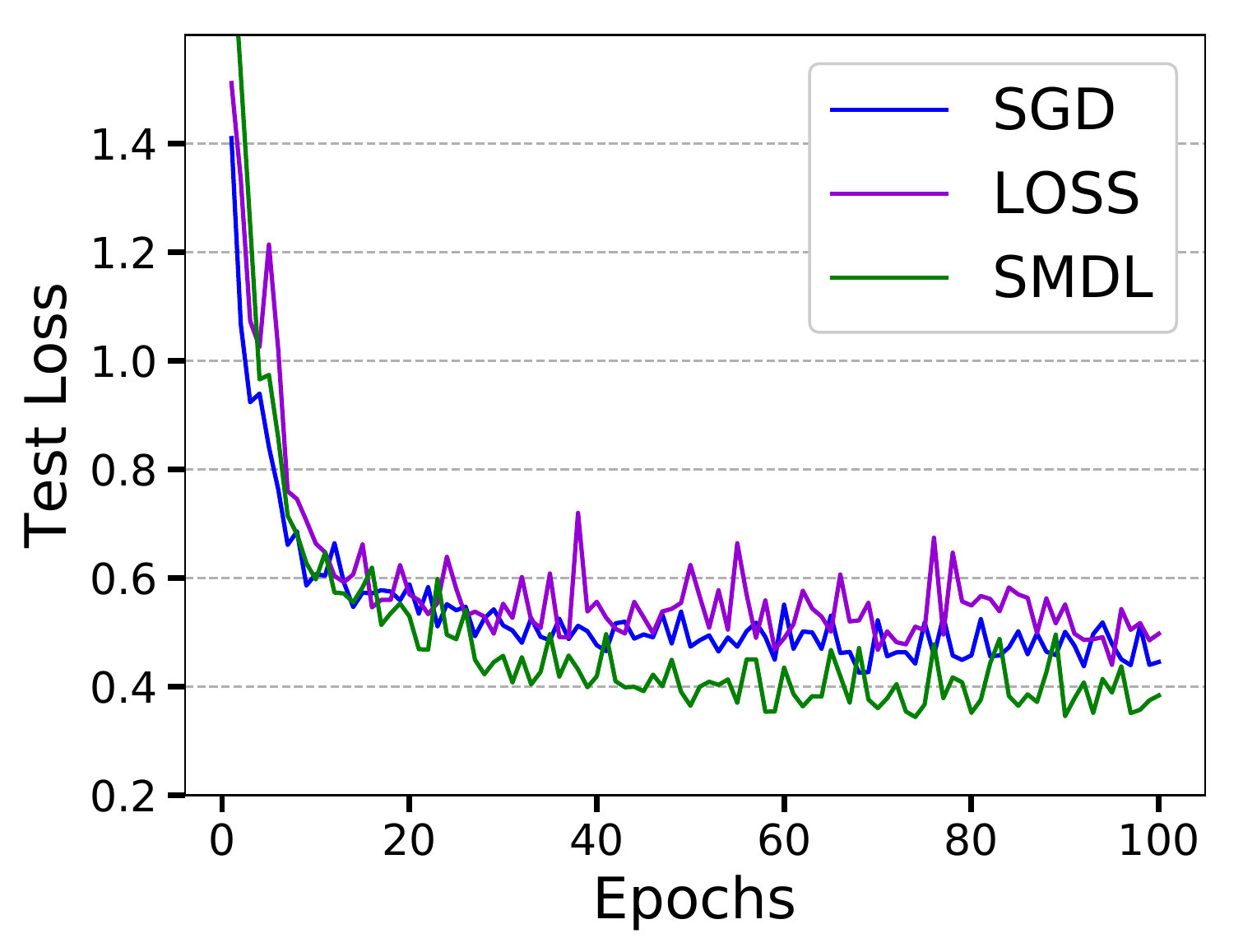}}
\hspace{15pt}\subfloat[Test Loss on CIFAR-100]{\includegraphics[width=0.30\linewidth]{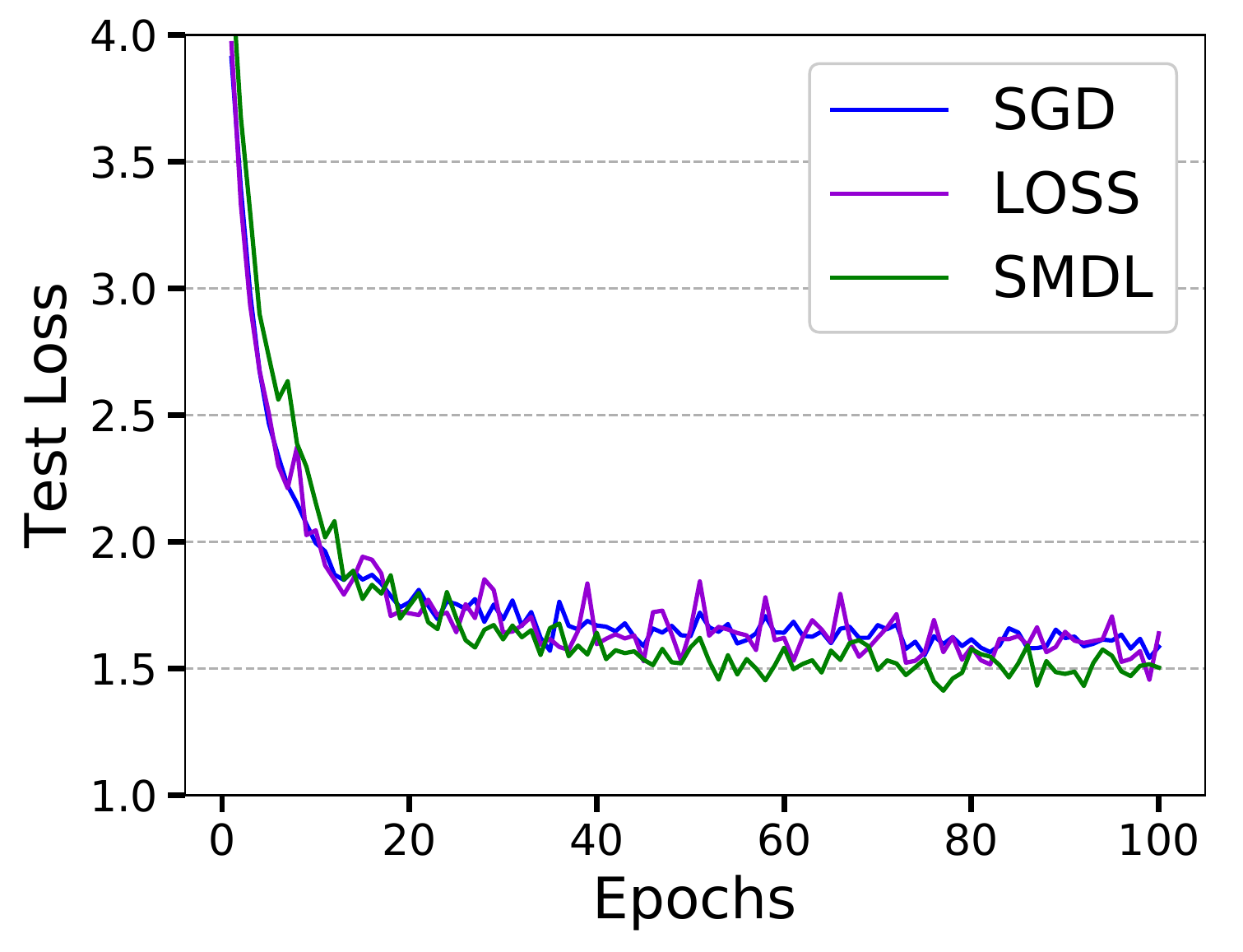}}
\vspace{-5pt}
\caption{Comparison of the proposed SMDL, with SGD and loss based sampling scheme on SVHN, CIFAR-10 and CIFAR-100 datasets. The test error and test loss is plotted. SMDL \textit{consistently outperforms} the baselines, both in terms of loss and generalization performance.}
\vspace{-10pt}
\label{fig:accuracy_error_plot}
\end{figure*}
%%%%%%%%% FIGURE ENDS

\section{Experiments and Results}\label{sec:experiments}

We conduct extensive experimental evaluations to study the effectiveness of submodular mini-batches in training deep neural networks over Stochastic Gradient Descent (SGD) and Loss-based sampling \cite{loshchilov2015online}, as in earlier efforts such as \cite{katharopoulos2018not}. For brevity, we refer to our proposed method of selecting submodular mini-batches for training as \textbf{S}ub\textbf{M}odular \textbf{D}ata \textbf{L}oader (\textbf{SMDL}).
% We report the results of those experiments in this section. 
We study the performance on the standard image classification task (as used in related earlier efforts) with SVHN \cite{netzer2011reading}, CIFAR-10 and CIFAR-100 \cite{krizhevsky2009learning} datasets. ResNet 20 \cite{he2016deep} is used as the network architecture for SVHN and CIFAR-10, while ResNet 32 is used with CIFAR-100.
Our implementation details are described in Sec \ref{sec:r_implementation_details}, followed by the main result and various ablation study results in Sec \ref{sec:r_learning}.

\subsection{Implementation Details}
\label{sec:r_implementation_details}
Algorithm \ref{algo:Training Algorithm} gives the generalized training procedure for submodular \mb selection. In each iteration, the current model $w_\tau^j$ is used to evaluate the submodular objective score (Equation \ref{eqn:full}). The feature representation for the images is obtained from the penultimate fully connected layer of this model. 
The probability values ($P(y|x_i, w)$) that are used in the computation of \textit{Uncertainty Score} is the softmax output from the model. 
Euclidean distance between the image features is used for \textit{Redundancy Score} computation in Equation \ref{eqn:Redundancy_Score}. We do an ablation study on the effect of using other distance metrics in Section \ref{sec:r_distance_metrics}. 
\textit{Mean Closeness Score} is computed as the cosine similarity between each data-point and the mean of all the training examples.

We follow the method used in \cite{brahma2018subset} and \cite{zhou2018minimax} to compute the fixed feature set $U$ used for evaluating \textit{Feature Match Score} (Equation \ref{eqn:feature_match}). We train a corresponding neural network, say $M$, on a random subset of training data, for an epoch. The features from the penultimate fully connected layer of $M$ is used as $U$. Square root function is used as $g(.)$. $m_u(x_i)$ is the feature at $u^{th}$ index of the representation of $x_i$ from the model $M$.

After a grid search and an empirical study, we use the following values for the co-efficients of the terms in the objective function: $\lambda_1 = \text{0.2}, \lambda_2 = \text{0.1}, \lambda_3 = \text{0.5}, \lambda_4 = \text{0.2}$. Ablation studies of the effect of the $\lambda$ parameters are presented in Section \ref{sec:r_trade_off_param}. Each of the scores is individually normalized across the selected pool of samples, before being combined, to ensure fair contribution. 
All the score computation (which depends on the softmax output and the feature representation from fully-connected layers) is a function of the model at each iteration (Line 5, Algorithm \ref{algo:Training Algorithm}). As we are using gradient descent for updating the parameters, we know that the model does not change drastically between iterations. Computational efficiency can be improved if we share the same model between successive iterations. 
We use a \textit{refresh rate} of 5 for all the experiments. A study of how our method behaves with different refresh rates is shown in Figure \ref{fig:ablation}(d). 
We note that increasing the refresh rate decreases the performance of the model. This is because the model changes over multiple iterations, which in turn affects the quality of the mini-batch selected.
% , which follows from the consensus that the model changes itself 
% We experiment with different \textit{refresh rates} and find that a refresh rate of $5$ would be ideal.
% Thus, the same model would be shared between five successive iterations.

We develop a modularized and configuration-driven tool in PyTorch \cite{paszke2017automatic}, which implements submodular selection and the other two baseline methods: SGD and Loss based sampling \cite{loshchilov2015online}. All the experiments are run for 100 epochs with a batch size of 50, a momentum parameter of $0.9$ and weight decay of $0.0001$. 
Use of batch normalization \cite{ioffe2015batch} and adaptive learning methods like Adam \cite{kingma2014adam} will complement the reported results of SMDL and other baseline methods. 
For SGD, all the reported results are the average of five runs. The partition size ($m$ in Algorithm \ref{algo:Sampling Algorithm} and \ref{algo:Training Algorithm}) is set to 10. Code is open-sourced: https://josephkj.in/projects/SMDL
\begin{table}[]
\resizebox{0.5\textwidth}{!}{%
\begin{tabular}{@{}llccccccccc@{}}
\toprule
Dataset                   &       & \multicolumn{3}{c}{SVHN} & \multicolumn{3}{c}{CIFAR-10} & \multicolumn{3}{c}{CIFAR-100} \\
Method                    &       & Loss Based  & SGD & SMDL & Loss Based   & SGD   & SMDL  & Loss Based   & SGD   & SMDL   \\ \midrule
\multirow{2}{*}{Accuracy(\%)} & Mean  &     90.87   & 93.60    &  \textbf{95.46}  &     81.06         &  82.54     &  \textbf{84.58}     &  53.77            &   53.57    &  \textbf{57.23}      \\
                          & Final &     92.44   & 94.34    &  \textbf{95.49}  &     84.32         &  85.63     &    \textbf{87.76}   &   57.22            & 58.27      &  \textbf{62.95}        \\
\multirow{2}{*}{Loss}     & Mean  &     0.363   & 0.230     & \textbf{0.175}  &     0.590         &  0.535     &     \textbf{0.487}  &   1.755            & 1.764       &   \textbf{1.717}     \\
                          & Final &     0.318  & 0.215    &    \textbf{0.182}   &   0.497    & 0.445      & \textbf{0.384}      &        1.639      &   1.586    &      \textbf{1.504}  \\ \bottomrule
\end{tabular}%
}
\vspace{-10pt}
\caption{Quantitative comparison of the proposed SMDL with SGD and Loss based sampling scheme on SVHN, CIFAR-10 and CIFAR-100 datasets. \textit{Mean} refers to the mean accuracy(\%) across epochs and \textit{Final} refers to the final accuracy.}
\vspace{-15pt}
\label{table:comparision_between_methods}
\end{table}
% -- Table_new_across_dataset ends -- 
\subsection{Results}\label{sec:r_learning}
We present the major result of our proposed submodular \mb selection method, SMDL in Figure \ref{fig:accuracy_error_plot} and Table \ref{table:comparision_between_methods}. We train the two network architectures on three datasets as enumerated in Section \ref{sec:experiments}. The generalization performance of these classification models, as measured by their test accuracy and test loss, is used as the evaluation metric. 
We see from Figure \ref{fig:accuracy_error_plot} and Table \ref{table:comparision_between_methods} that SMDL is able to achieve lower error and loss, \textit{consistently} across epochs, \textit{on all the three} datasets. 
It is worth noting that Loss-based sampling fails significantly on SVHN. Such deterioration of generalization performance is also noted in \cite{loshchilov2015online}.
% SMDL maintains its performance consistently across all the datasets. 
The values reported on SGD are its mean after conducting five trials.

These results support our claim that selecting a \mb which respects diversity and informativeness of the samples helps in more generalizable deep learning models.

% %%%%%%%%% FIGURE STARTS
% \begin{figure}
% \centering
% \subfloat[Test Error against epochs.]{\includegraphics[width=0.52\linewidth]{plots/SVHN_Test_Error__alphas_.pdf}}
% \subfloat[Test Loss against epochs.]{\includegraphics[width=0.52\linewidth]{plots/SVHN_Test_Loss__alphas_.pdf}}
% % \vspace{-12.6pt}
% \caption{Figure shows how each trade-off parameter influence the quality of the solution. 
% Each of the score function individually has better performance than SGD while a combination of them performs best.
% Refer Section \ref{sec:r_trade_off_param} for more details.
% }
% \vspace{-10pt}
% \label{fig:tradeoff_ablation}
% \end{figure}
% %%%%%%%%% FIGURE ENDS

% -- Table_lambda_values begins -- 
% \usepackage{booktabs}
% \usepackage{graphicx}
\begin{table}[]
\vspace{-15pt}
\resizebox{0.5\textwidth}{!}{%
\begin{tabular}{@{}ccccccccccc@{}}
\toprule
 & \multicolumn{2}{c}{0} & \multicolumn{2}{c}{0.2} & \multicolumn{2}{c}{0.5} & \multicolumn{2}{c}{0.8} & \multicolumn{2}{c}{1.0} \\
 & Mean & Final & Mean & Final & Mean & Final & Mean & Final & Mean & Final \\ \midrule
$\lambda_1$ & 64.41 & 87.19  & 58.60 & 91.13 & \textbf{66.99} & \textbf{92.39} & 65.68 & 92.27 & 65.80 & 92.12 \\
$\lambda_2$ & 46.17 & 83.62 & 64.55 & 91.34 & \textbf{68.28} & \textbf{92.33} & 67.70 & 90.77 & 61.34 & 85.06 \\
$\lambda_3$ & 64.41 & 87.19 & 62.90 & 91.01 & 60.91 & 89.50 & \textbf{63.58} & \textbf{91.02} & 59.06 & 89.77 \\
$\lambda_4$ &  64.41 & 87.19 & 62.25 & \textbf{92.14} & 63.03 & 86.90 & \textbf{65.50} & 91.03 & 60.30 & 86.61 \\ \bottomrule
\end{tabular}%
}
\vspace{-5pt}
\caption{Ablation study of $\lambda$ parameters on subset of SVHN dataset. \textit{Mean} across epochs and accuracy(\%) of the \textit{Final} model is reported. 
% We use Final Accuracy to draw comparison and Mean Accuracy is reported to get the sense of the overall performance across epochs.
}
\vspace{-5pt}
\label{table:Alpha_Ablation}
\end{table}
% -- Table_lambda_values ends -- 

\subsection{Ablation Studies}
\paragraph{Effect of trade-off parameters.}\label{sec:r_trade_off_param}
%We study the effect of the four trade-off parameters that control the contribution of each of the four score functions in the submodular objective function (Equation \ref{eqn:full}) here. Table \ref{table:ablation_trade_off} summarises the result on SVHN dataset. 
% Figure \ref{fig:tradeoff_ablation} shows how test error and test loss changes over epochs.

We study the effect of trade-off parameters that control the contribution of each of the four score functions in the submodular objective function (Equation \ref{eqn:full}). 
For this, we train ResNet 20 on a small subset of SVHN dataset.
% Each $\lambda_{i}$ is picked from $\{0, 0.2, 0.5, 0.8, 1.0\}$.
Except for the Redundancy Score (controlled via $\lambda_2$), all the other terms are modular. Hence $\lambda_2$ should be non-zero to make the objective function submodular.
We vary each of the other $\lambda_{i} (i \in\{{1,3,4}\})$ with values from $\{0, 0.2, 0.5, 0.8, 1.0\}$, by fixing $\lambda_{2}$=0.5 and rest to zero. These results are reported in row 1, 3 and 4 of Table \ref{table:Alpha_Ablation}.
% We have done this to study the relative importance of modular terms (Equations \ref{eqn:Uncertainity_Score}, \ref{eqn:Mean_Closeness_Score}, \ref{eqn:feature_match})  w.r.t submodular term (Equation \ref{eqn:Redundancy_Score}) in our formulation. 
Then, we fix the best values obtained for $\lambda_{1}, \lambda_{3} \text{ and } \lambda_{4}$ and vary $\lambda_{2}$. The result is populated in the second row of Table \ref{table:Alpha_Ablation}.
% To study the effect of $\lambda_{2}$ we first identified the best performing $\lambda_{i} (i \in\{{1,3,4}\})$ followed by varying $\lambda_{2}$ after fixing $\lambda_{i} (i \in\{{1,3,4}\})$ with best values obtained.  Table \ref{table:Alpha_Ablation} summarises these results obtained on a subset of SVHN dataset.
It is evident from the table that the following trade-off parameters (after normalization) achieves best performance on the subset:
 $\lambda_1 = \text{0.25}, \lambda_2 = \text{0.25}, \lambda_3 = \text{0.4}, \lambda_4 = \text{0.1}$. 
%$\{ {0.25, 0.25, 0.4, 0.1} \}$ achieves best performance on the subset. 
Empirically, we find that 
 $\lambda_1 = \text{0.2}, \lambda_2 = \text{0.1}, \lambda_3 = \text{0.5}, \lambda_4 = \text{0.2}$, 
%$\{ {0.2, 0.1, 0.5, 0.2} \}$
achieves best performance on the whole dataset. These set of trade-off parameters also generalises well to CIFAR-10 and CIFAR-100 datasets.

%We train four models, SMDL-\{1,2,3,4\}, each with higher weightage to Uncertainity Score, Redundancy Score, Mean Closeness Score and Feature Match Score respectively to measure each of its influence. 
From Table \ref{table:Alpha_Ablation} we can observe that each of the score function has a profound effect on the quality of the model being trained. Setting each of them to zero hurts the performance the most (Column 2 and 3).
The second row reveals that the submodular term has the maximal impact if we set it zero.
Each of the scores is independently competent while combining them gives the best performance.
% This emphasizes the fact that every score function is important and the overall formulation (Equation \ref{eqn:full}) achieves best performance.

%The quality of all these models are better than the ones trained with SGD and loss based sampling \cite{loshchilov2015online}.
%Model with higher weightage for Redundancy Score and Mean Closeness Score has slightly better accuracy and lower loss. These scores encourages diversity in the subset, while keeping outliers away. 
%We slightly increase the weightage to these scores to finalize our final submodular objective function (last row of Table \ref{table:ablation_trade_off}). This is the model that is used to report all the other results in the paper.

%%%%%%%%% FIGURE STARTS
\begin{figure}
\centering
\vspace{-29pt}
\subfloat[Comparison with different distance metrics.]{\includegraphics[width=0.49\linewidth]{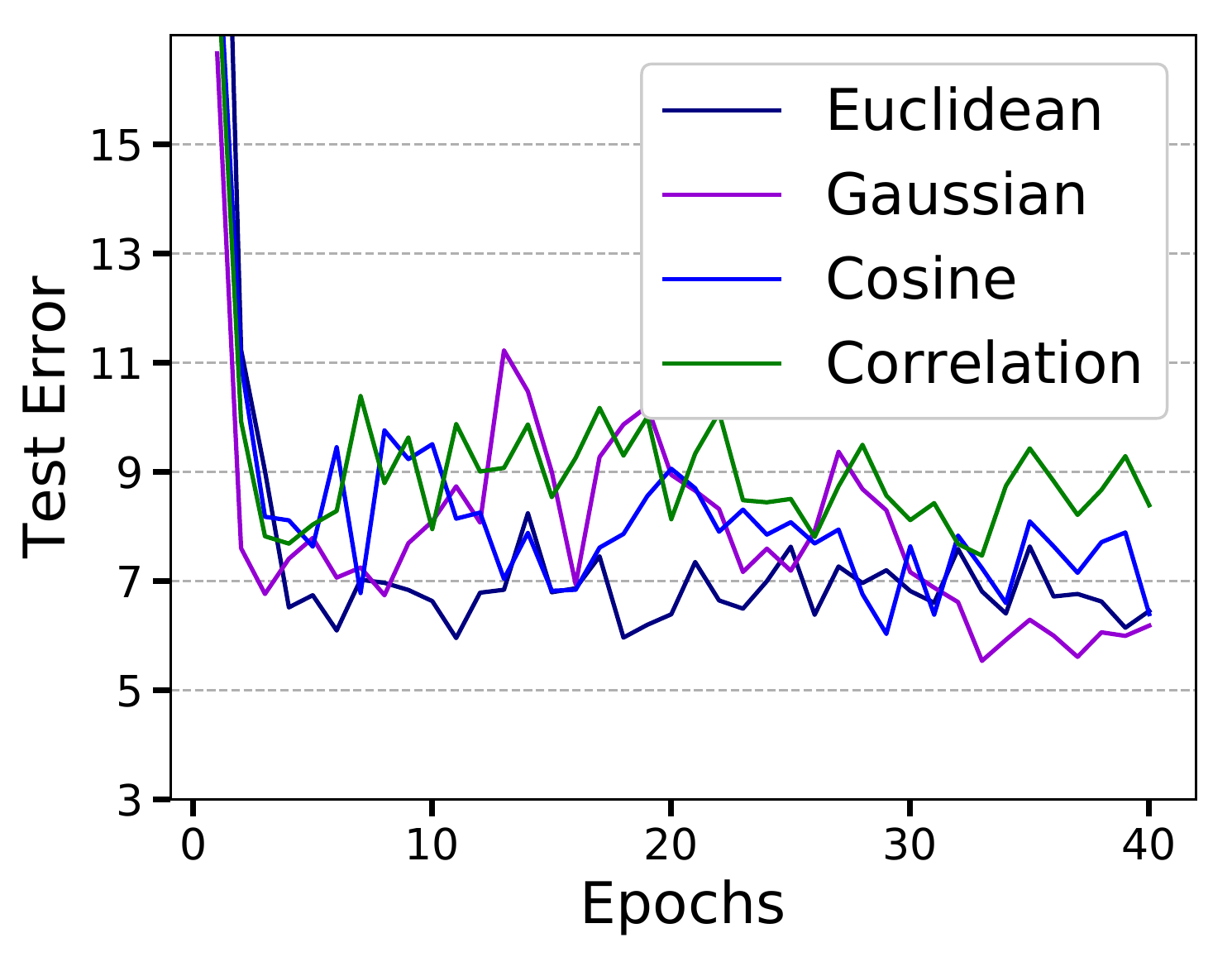}}
\hspace{3pt}\subfloat[Test Error plot with different learning rates (LR).]{\includegraphics[width=0.49\linewidth]{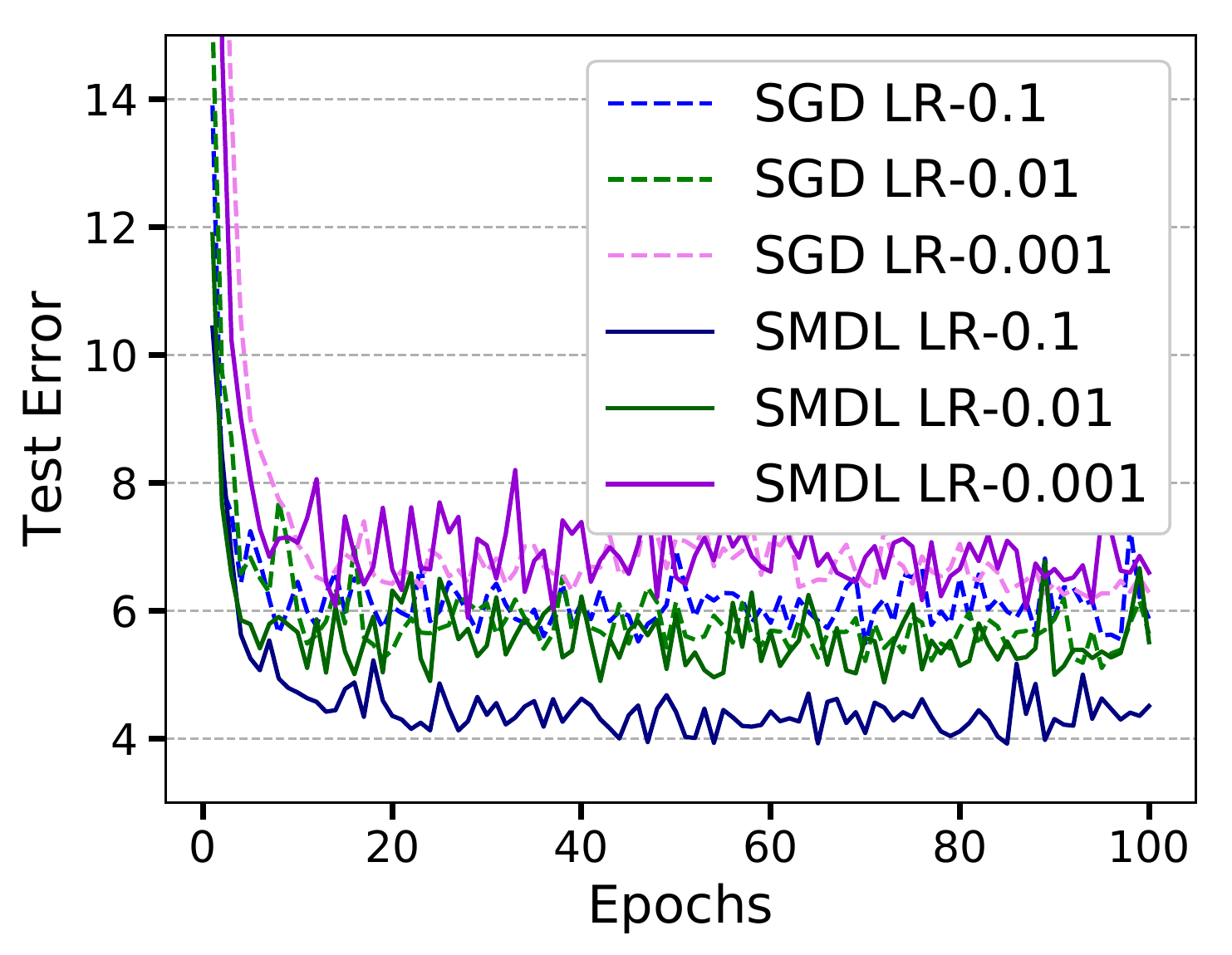}}
\vspace{-10pt}
\subfloat[Test Error plot with different mini-batch sizes (BS).]{\includegraphics[width=0.49\linewidth]{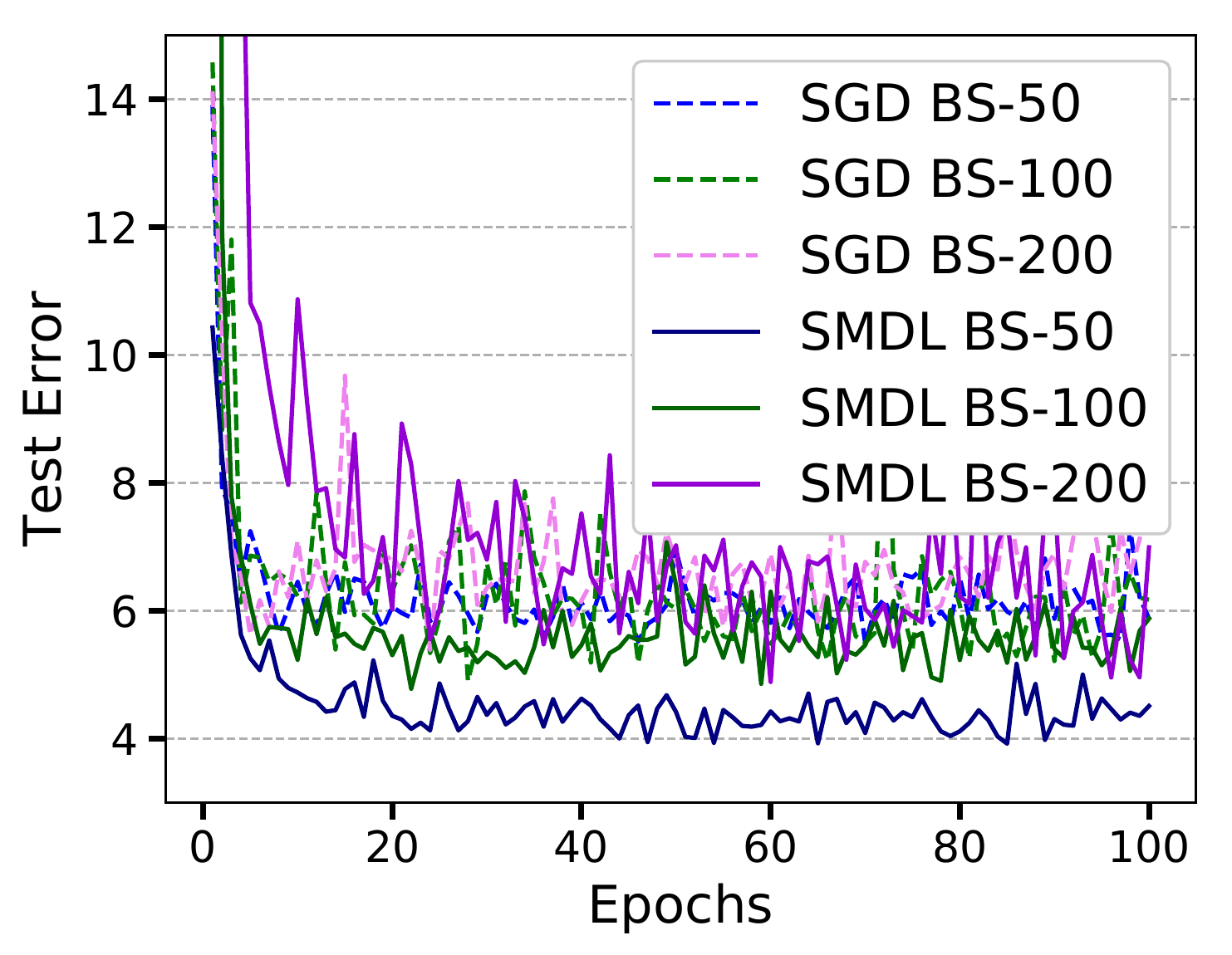}}
\hspace{3pt}\subfloat[Test Error plot with different Refresh Rates (RR).]{\includegraphics[width=0.49\linewidth]{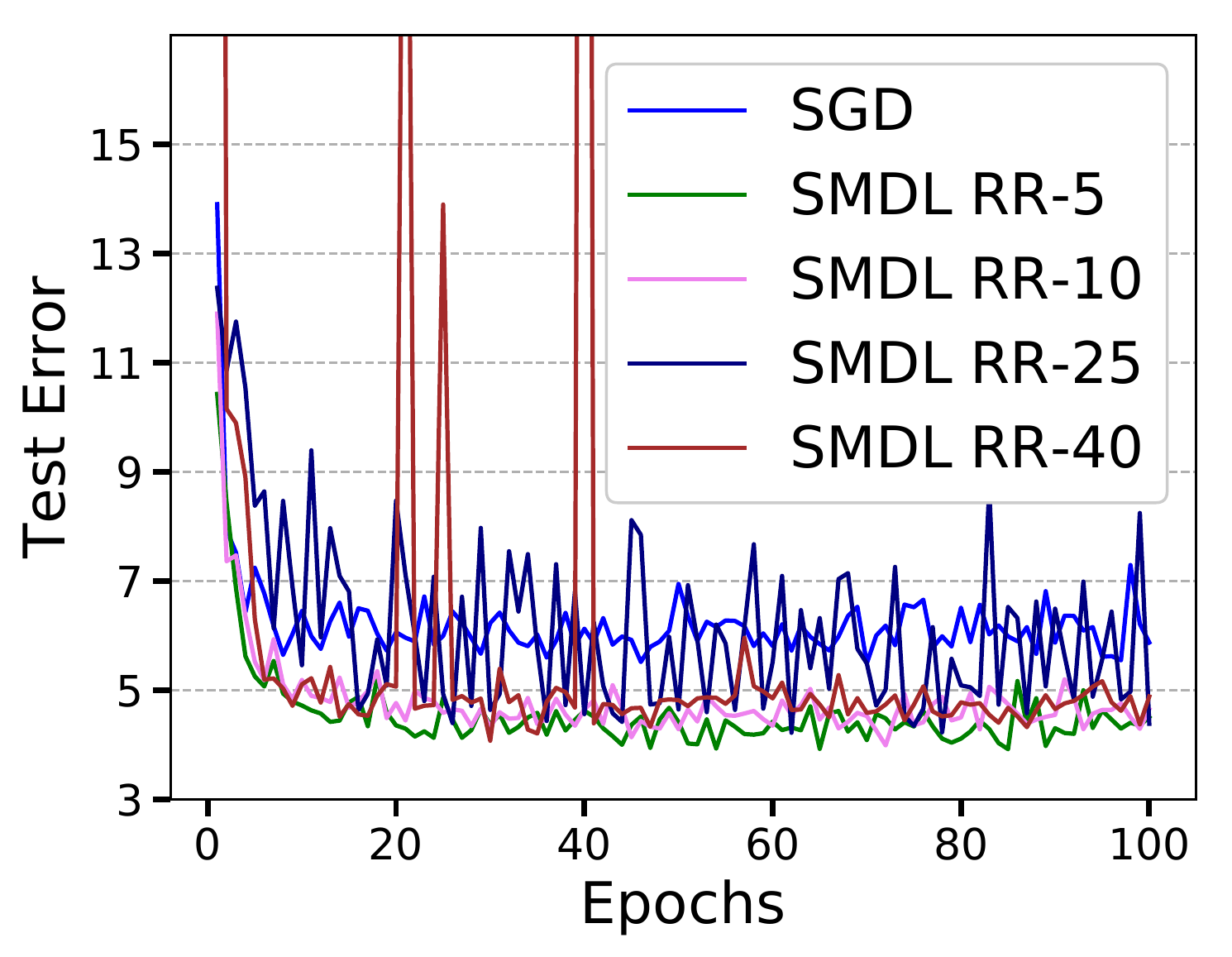}}
\vspace{-2pt}
\caption{The figure shows the ablation results on SVHN dataset.
1)We find that the Euclidean distance measure performs best among other distance metrics. 2)The proposed method SMDL consistently outperforms SGD even with different batch sizes and different learning rates. We note that a batch size of $50$ and a learning rate of $0.1$ gives the least error for SMDL. 3) We find that a refresh rate of 5 gives the best performance. (refer Section \ref{sec:r_implementation_details}).}
\label{fig:ablation}
\vspace{-12pt}
\end{figure}
%%%%%%%%% FIGURE ENDS

\paragraph{Variations across different distance metrics}\label{sec:r_distance_metrics}
We perform an ablation study on the impact of different distance metrics while computing the value of the submodular objective function.
We choose four distance metrics and evaluate its impact on training an image classifier on SVHN dataset.
Assuming $u$ and $v$ as the two vectors, the distance metrics considered are: Euclidean ($\lVert u - v\rVert_2$), Cosine ($1 - \frac{u\cdot v}{\lVert u \rVert_2 \lVert v \rVert_2}$), Correlation ($1 - \frac{(u - \bar{u})\cdot (v - \bar{v})}{\lVert (u - \bar{u}) \rVert_2 \lVert (v - \bar{v}) \rVert_2}$) where $\bar{u}$ is the mean elements of vector $u$ and Gaussian ($\exp{-\frac{(\lVert u - v\rVert^2)}{2 \sigma^2}}$). 

Figure \ref{fig:ablation}(a) shows the result of the experiment, where the test error is plotted against epochs. These results suggest that the Euclidean distance metric gives lower test error than the others. 

\paragraph{Effect of learning rate and batch size}\label{sec:r_lr_batch_size}
Batch size and learning rate are the most important hyper-parameters that impact the learning dynamics of the model. In order to study the robustness of the model trained with submodular mini-batches, we vary batch size and learning rate, keeping all other parameters the same. 
We compare with SGD trained with the same set of hyper-parameters for a fair comparison.

Figure \ref{fig:ablation}(c) reports the results when mini-batch sizes are set to 50, 100 and 200. Our results agree with the common consensus \cite{li2014efficient} that increasing the batch-size decreases the rate of convergence, still SMDL beats SGD by a consistent margin (solid lines in the graph) for all the batch sizes. We set the learning rate to 0.1, 0.01 and 0.001 and observe the same similar pattern in Figure \ref{fig:ablation}(b). These results show that the performance of SMDL is robust to learning rate and batch size changes. 

\vspace{-5pt}
\section{Conclusion}\label{sec:conclusion}
In this work, we cast the selection of diverse and informative mini-batches for training a deep learning model as a submodular optimization problem. We design a novel submodular objective and propose a scalable algorithm to do submodular selection. 
Extensive experimental valuation on three datasets reveals significant improvement in convergence and generalization performance of the model trained with submodular mini-batches over SGD and Loss based sampling \cite{loshchilov2015online}.
The ablation results show that the method is robust to changes in batch size and learning rate. 

It would be ideal to give the model the expressive power to decide not just which data points to select, but also the number of such data points to train on. 
This can be easily done by incorporating an additional factor into the submodular objective function, which checks the growth of the mini-batch size. This would be an important direction of our future work. 
In addition, more efficient strategies to speed up the proposed method further would form a key interest of ours in our future efforts.
%We would explore this as the immediate next step. We also note that the submodular optimization framework that we have come up with can be easily adapted for many other problems where subset selection is the major task involved. This can find application in selecting those samples to be labeled in an active learning setup and those exemplars to be stored in the exemplar memory in a class incremental learning setup. A carefully designed submodular objective function which can guarantee diverse subsets and an efficient implementation strategy would be invaluable for those settings.

% \section*{Acknowledgement}
\paragraph{Acknowledgements}
We would like to thank MHRD, Govt of India for funding the personnel involved, Microsoft Research India and ACM-India/IARCS for the travel grants to present the work and the anonymous reviewers for their valuable feedback.

\clearpage
\bibliographystyle{named}
{\small
\bibliography{ijcai19}}

\begin{thebibliography}{}

\bibitem[\protect\citeauthoryear{Alain \bgroup \em et al.\egroup
  }{2015}]{alain2015variance}
Guillaume Alain, Alex Lamb, Chinnadhurai Sankar, Aaron Courville, and Yoshua
  Bengio.
\newblock Variance reduction in sgd by distributed importance sampling.
\newblock {\em arXiv preprint arXiv:1511.06481}, 2015.

\bibitem[\protect\citeauthoryear{Allen-Zhu}{2017}]{allen2017katyusha}
Zeyuan Allen-Zhu.
\newblock Katyusha: The first direct acceleration of stochastic gradient
  methods.
\newblock {\em The Journal of Machine Learning Research}, 18(1):8194--8244,
  2017.

\bibitem[\protect\citeauthoryear{Brahma and Othon}{2018}]{brahma2018subset}
Pratik~Prabhanjan Brahma and Adrienne Othon.
\newblock Subset replay based continual learning for scalable improvement of
  autonomous systems.
\newblock In {\em CVPR Workshops}, pages 1179--11798. IEEE, 2018.

\bibitem[\protect\citeauthoryear{Chakraborty \bgroup \em et al.\egroup
  }{2015}]{chakraborty2015adaptive}
Shayok Chakraborty, Vineeth Balasubramanian, and Sethuraman Panchanathan.
\newblock Adaptive batch mode active learning.
\newblock {\em IEEE TNNLS}, 26(8):1747--1760, 2015.

\bibitem[\protect\citeauthoryear{Chang \bgroup \em et al.\egroup
  }{2017}]{chang2017active}
Haw-Shiuan Chang, Erik Learned-Miller, and Andrew McCallum.
\newblock Active bias: Training more accurate neural networks by emphasizing
  high variance samples.
\newblock In {\em Advances in Neural Information Processing Systems}, pages
  1002--1012, 2017.

\bibitem[\protect\citeauthoryear{Das and Kempe}{2008}]{das2008algorithms}
Abhimanyu Das and David Kempe.
\newblock Algorithms for subset selection in linear regression.
\newblock In {\em ACM STOC}, pages 45--54. ACM, 2008.

\bibitem[\protect\citeauthoryear{He \bgroup \em et al.\egroup
  }{2016}]{he2016deep}
Kaiming He, Xiangyu Zhang, Shaoqing Ren, and Jian Sun.
\newblock Deep residual learning for image recognition.
\newblock In {\em CVPR}, pages 770--778, 2016.

\bibitem[\protect\citeauthoryear{Ioffe and Szegedy}{2015}]{ioffe2015batch}
Sergey Ioffe and Christian Szegedy.
\newblock Batch normalization: Accelerating deep network training by reducing
  internal covariate shift.
\newblock {\em arXiv preprint arXiv:1502.03167}, 2015.

\bibitem[\protect\citeauthoryear{Johnson and
  Zhang}{2013}]{johnson2013accelerating}
Rie Johnson and Tong Zhang.
\newblock Accelerating stochastic gradient descent using predictive variance
  reduction.
\newblock In {\em NIPS}, pages 315--323, 2013.

\bibitem[\protect\citeauthoryear{Katharopoulos and
  Fleuret}{2017}]{katharopoulos2017biased}
Angelos Katharopoulos and Fran{\c{c}}ois Fleuret.
\newblock Biased importance sampling for deep neural network training.
\newblock {\em arXiv preprint arXiv:1706.00043}, 2017.

\bibitem[\protect\citeauthoryear{Katharopoulos and
  Fleuret}{2018}]{katharopoulos2018not}
Angelos Katharopoulos and Fran{\c{c}}ois Fleuret.
\newblock Not all samples are created equal: Deep learning with importance
  sampling.
\newblock {\em arXiv:1803.00942}, 2018.

\bibitem[\protect\citeauthoryear{Kingma and Ba}{2014}]{kingma2014adam}
Diederik~P Kingma and Jimmy Ba.
\newblock Adam: A method for stochastic optimization.
\newblock {\em arXiv preprint arXiv:1412.6980}, 2014.

\bibitem[\protect\citeauthoryear{Krizhevsky and
  Hinton}{2009}]{krizhevsky2009learning}
Alex Krizhevsky and Geoffrey Hinton.
\newblock Learning multiple layers of features from tiny images.
\newblock Technical report, Citeseer, 2009.

\bibitem[\protect\citeauthoryear{Li \bgroup \em et al.\egroup
  }{2014}]{li2014efficient}
Mu~Li, Tong Zhang, Yuqiang Chen, and Alexander~J Smola.
\newblock Efficient mini-batch training for stochastic optimization.
\newblock In {\em Proceedings of the 20th ACM SIGKDD international conference
  on Knowledge discovery and data mining}, pages 661--670. ACM, 2014.

\bibitem[\protect\citeauthoryear{Li \bgroup \em et al.\egroup
  }{2016}]{li2016fast}
Chengtao Li, Stefanie Jegelka, and Suvrit Sra.
\newblock Fast dpp sampling for nystr$\backslash$" om with application to
  kernel methods.
\newblock {\em arXiv preprint arXiv:1603.06052}, 2016.

\bibitem[\protect\citeauthoryear{Lin and Bilmes}{2011}]{lin2011class}
Hui Lin and Jeff Bilmes.
\newblock A class of submodular functions for document summarization.
\newblock In {\em ACL}, pages 510--520. Association for Computational
  Linguistics, 2011.

\bibitem[\protect\citeauthoryear{Loshchilov and
  Hutter}{2015}]{loshchilov2015online}
Ilya Loshchilov and Frank Hutter.
\newblock Online batch selection for faster training of neural networks.
\newblock {\em ICLR Workshops}, 2015.

\bibitem[\protect\citeauthoryear{Minoux}{1978}]{minoux1978accelerated}
Michel Minoux.
\newblock Accelerated greedy algorithms for maximizing submodular set
  functions.
\newblock In {\em Optimization techniques}, pages 234--243. Springer, 1978.

\bibitem[\protect\citeauthoryear{Mirzasoleiman \bgroup \em et al.\egroup
  }{2013}]{mirzasoleiman2013distributed}
Baharan Mirzasoleiman, Amin Karbasi, Rik Sarkar, and Andreas Krause.
\newblock Distributed submodular maximization: Identifying representative
  elements in massive data.
\newblock In {\em Advances in Neural Information Processing Systems}, pages
  2049--2057, 2013.

\bibitem[\protect\citeauthoryear{Mirzasoleiman \bgroup \em et al.\egroup
  }{2015}]{mirzasoleiman2015lazier}
Baharan Mirzasoleiman, Ashwinkumar Badanidiyuru, Amin Karbasi, Jan Vondr{\'a}k,
  and Andreas Krause.
\newblock Lazier than lazy greedy.
\newblock In {\em AAAI}, 2015.

\bibitem[\protect\citeauthoryear{Nemhauser \bgroup \em et al.\egroup
  }{1978}]{nemhauser1978analysis}
George~L Nemhauser, Laurence~A Wolsey, and Marshall~L Fisher.
\newblock An analysis of approximations for maximizing submodular set
  functions—i.
\newblock {\em Mathematical programming}, 14(1):265--294, 1978.

\bibitem[\protect\citeauthoryear{Netzer \bgroup \em et al.\egroup
  }{2011}]{netzer2011reading}
Yuval Netzer, Tao Wang, Adam Coates, Alessandro Bissacco, Bo~Wu, and Andrew~Y
  Ng.
\newblock Reading digits in natural images with unsupervised feature learning.
\newblock In {\em NIPS workshop}, page~5, 2011.

\bibitem[\protect\citeauthoryear{Paszke \bgroup \em et al.\egroup
  }{2017}]{paszke2017automatic}
Adam Paszke, Sam Gross, Soumith Chintala, Gregory Chanan, Edward Yang, Zachary
  DeVito, Zeming Lin, Alban Desmaison, Luca Antiga, and Adam Lerer.
\newblock Automatic differentiation in pytorch.
\newblock In {\em NIPS-W}, 2017.

\bibitem[\protect\citeauthoryear{Shamaiah \bgroup \em et al.\egroup
  }{2010}]{shamaiah2010greedy}
Manohar Shamaiah, Siddhartha Banerjee, and Haris Vikalo.
\newblock Greedy sensor selection: Leveraging submodularity.
\newblock In {\em 49th IEEE conference on decision and control (CDC)}, pages
  2572--2577. IEEE, 2010.

\bibitem[\protect\citeauthoryear{Singh and
  Balasubramanian}{2018}]{singh2018submodular}
Krishna~Kant Singh and Vineeth~N Balasubramanian.
\newblock Submodular importance sampling for neural network training.
\newblock Master's thesis, Indian Institute of Technology Hyderabad, 2018.

\bibitem[\protect\citeauthoryear{Thangarasa and
  Taylor}{2018}]{thangarasa2018self}
Vithursan Thangarasa and Graham~W Taylor.
\newblock Self-paced learning with adaptive deep visual embeddings.
\newblock {\em arXiv preprint arXiv:1807.09200}, 2018.

\bibitem[\protect\citeauthoryear{Wei \bgroup \em et al.\egroup
  }{2014}]{wei2014submodular}
Kai Wei, Yuzong Liu, Katrin Kirchhoff, Chris Bartels, and Jeff Bilmes.
\newblock Submodular subset selection for large-scale speech training data.
\newblock In {\em ICASSP}, pages 3311--3315. IEEE, 2014.

\bibitem[\protect\citeauthoryear{Wei \bgroup \em et al.\egroup
  }{2015}]{wei2015submodularity}
Kai Wei, Rishabh Iyer, and Jeff Bilmes.
\newblock Submodularity in data subset selection and active learning.
\newblock In {\em ICML}, pages 1954--1963, 2015.

\bibitem[\protect\citeauthoryear{Zhang \bgroup \em et al.\egroup
  }{2017}]{zhang2017determinantal}
Cheng Zhang, Hedvig Kjellstrom, and Stephan Mandt.
\newblock Determinantal point processes for mini-batch diversification.
\newblock {\em arXiv preprint arXiv:1705.00607}, 2017.

\bibitem[\protect\citeauthoryear{Zhang \bgroup \em et al.\egroup
  }{2018}]{zhang2018active}
Cheng Zhang, Cengiz {\"O}ztireli, Stephan Mandt, and Giampiero Salvi.
\newblock Active mini-batch sampling using repulsive point processes.
\newblock {\em arXiv:1804.02772}, 2018.

\bibitem[\protect\citeauthoryear{Zhao and Zhang}{2014}]{zhao2014accelerating}
Peilin Zhao and Tong Zhang.
\newblock Accelerating minibatch stochastic gradient descent using stratified
  sampling.
\newblock {\em arXiv preprint arXiv:1405.3080}, 2014.

\bibitem[\protect\citeauthoryear{Zhao and Zhang}{2015}]{zhao2015stochastic}
Peilin Zhao and Tong Zhang.
\newblock Stochastic optimization with importance sampling for regularized loss
  minimization.
\newblock In {\em ICML}, pages 1--9, 2015.

\bibitem[\protect\citeauthoryear{Zhou and Bilmes}{2018}]{zhou2018minimax}
Tianyi Zhou and Jeff Bilmes.
\newblock Minimax curriculum learning: Machine teaching with desirable
  difficulties and scheduled diversity.
\newblock In {\em ICLR}, 2018.

\end{thebibliography}

\clearpage
\appendix
\begin{center}
\Large
    \textbf{Supplementary Section}
\end{center}

\section{Analysis of the computational complexity}

The asymptotic complexity of selecting each minibatch using Algorithm \ref{algo:Sampling Algorithm} is $m \times r^2 \times d$, where $m$ is the number of partition to which the training data is split into (Line 2 of Algorithm \ref{algo:Sampling Algorithm}), $r$ is the sample size of each of the Lazier than Lazy selection \cite{mirzasoleiman2015lazier} and $d$ is the dimension of the feature vector.

The divide and conquer strategy, along with the parallelism that can be achieved makes the proposed approach practically viable. The dataset $V$, is partitioned into $m$ random samples in line 2 of Algorithm \ref{algo:Sampling Algorithm}. Lazier than Lazy selection is run in parallel on multiple cores for each partition $V_i$ to pick $b$ samples (lines 5-8 of Algorithm \ref{algo:Sampling Algorithm}). These $b\times m$ samples are then combined and $b$ items are selected as mini-batch items (lines 9-14 of Algorithm \ref{algo:Sampling Algorithm}).

% \usepackage{booktabs}
\begin{table}[h]
\centering
\begin{tabular}{@{}cccc@{}}
\toprule
Datasets & \multicolumn{3}{c}{Methods} \\ \midrule
 & SMDL & Loss & SGD \\ \midrule
SVHN & 937.2305s & 7007.7064s & 373.1943s \\
CIFAR 10 & 948.7460s & 5196.3846s & 142.5396s \\
CIFAR 100 & 764.6001s & 8221.2820s & 160.0396s \\ \bottomrule
\end{tabular}
\caption{Comparison of the average time for completing one epoch (in seconds) taken by SMDL against Loss based sampling and SGD. }
\label{tab:time_comparison}
\end{table}

Table \ref{tab:time_comparison} compares average time taken for completing one epoch by SMDL, Loss based sampling \cite{loshchilov2015online} and SGD. It is evident that SMDL takes much more time than SGD but achieves better generalization capability and is much faster than other methods that accomplish the same task like Loss based sampling.

%%%%%%%%% FIGURE STARTS
\begin{figure}[h]
\centering
\vspace{-10pt}
\subfloat[Test Error plotted across epochs.]{\includegraphics[width=0.49\linewidth]{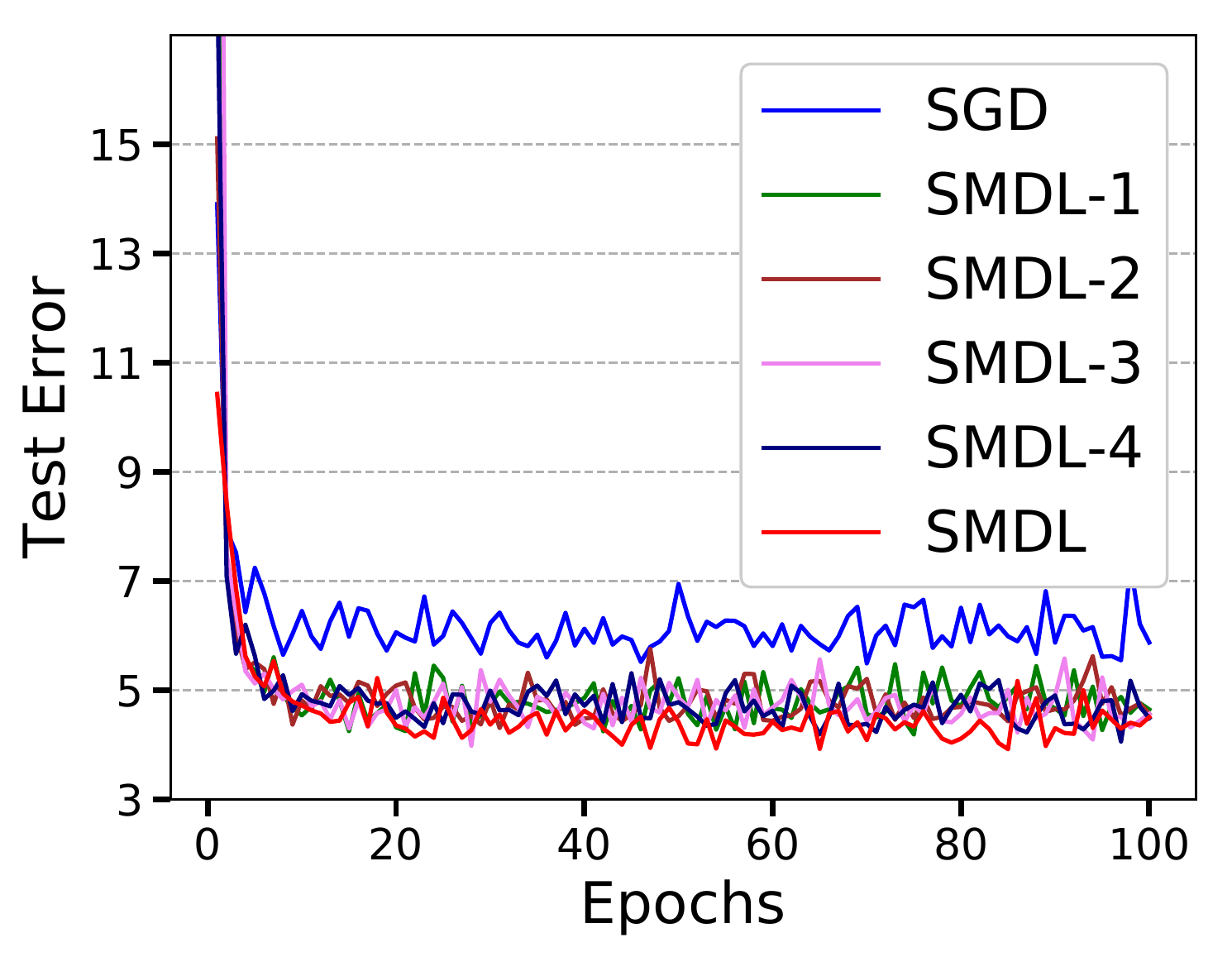}}
\hspace{3pt}\subfloat[Test Loss plotted across epochs.]{\includegraphics[width=0.49\linewidth]{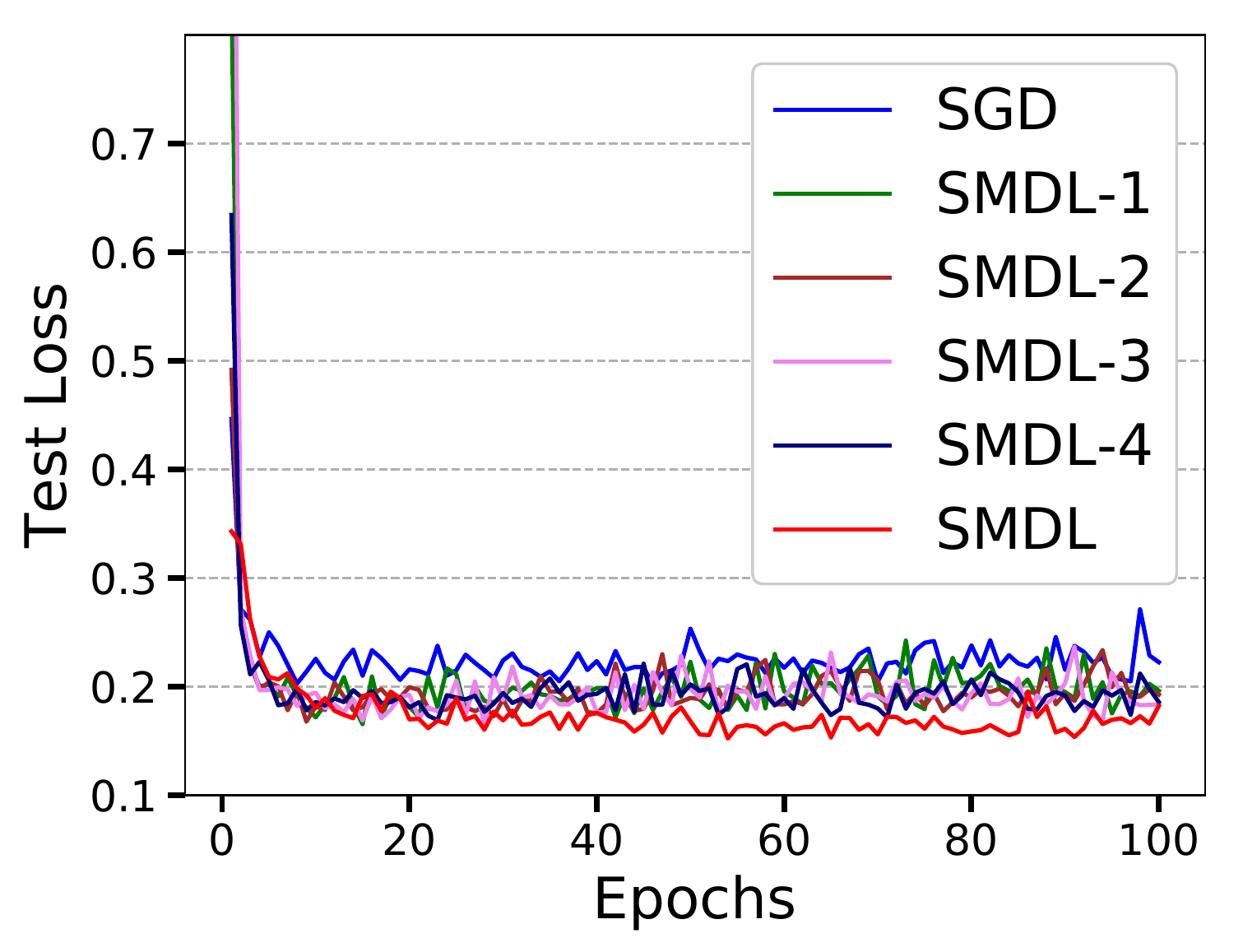}}
\vspace{-2pt}
\caption{The graph brings out the importance of each of the term in the submodular formulation (Equation \ref{eqn:full}). SMDL-\{1,2,3,4\} are four models which has only one of the trade-off parameter turned on. It is compared against SGD and standard SMDL.}
\label{fig:ablation_2}
\vspace{-6pt}
\end{figure}
%%%%%%%%% FIGURE ENDS

%%%%%%%%% FIGURE STARTS
\begin{figure}
\centering
\vspace{-27pt}
\subfloat[Comparison with different distance metrics.]{\includegraphics[width=0.49\linewidth]{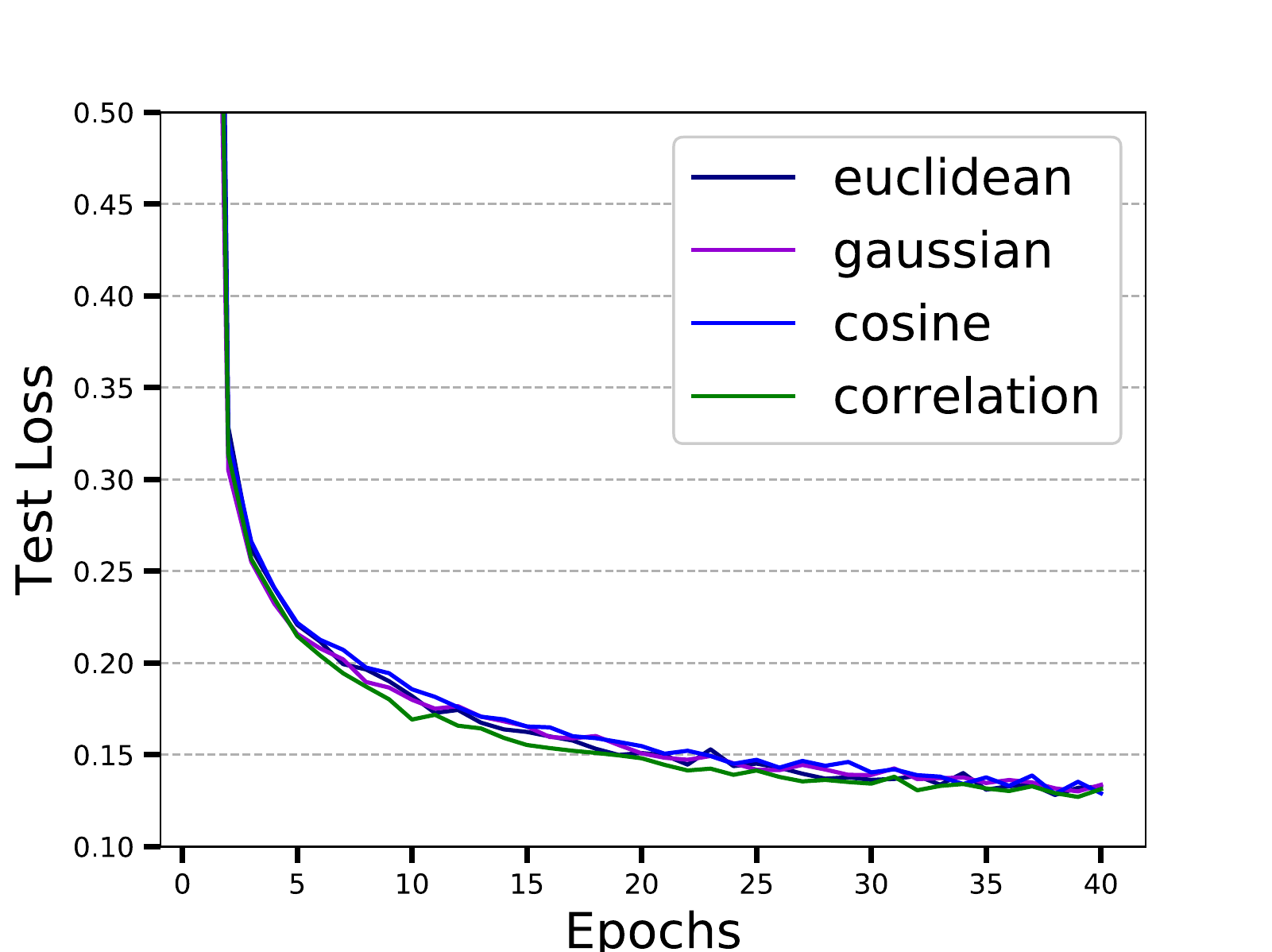}}
\hspace{3pt}\subfloat[Test Loss plot with different learning rates (LR).]{\includegraphics[width=0.49\linewidth]{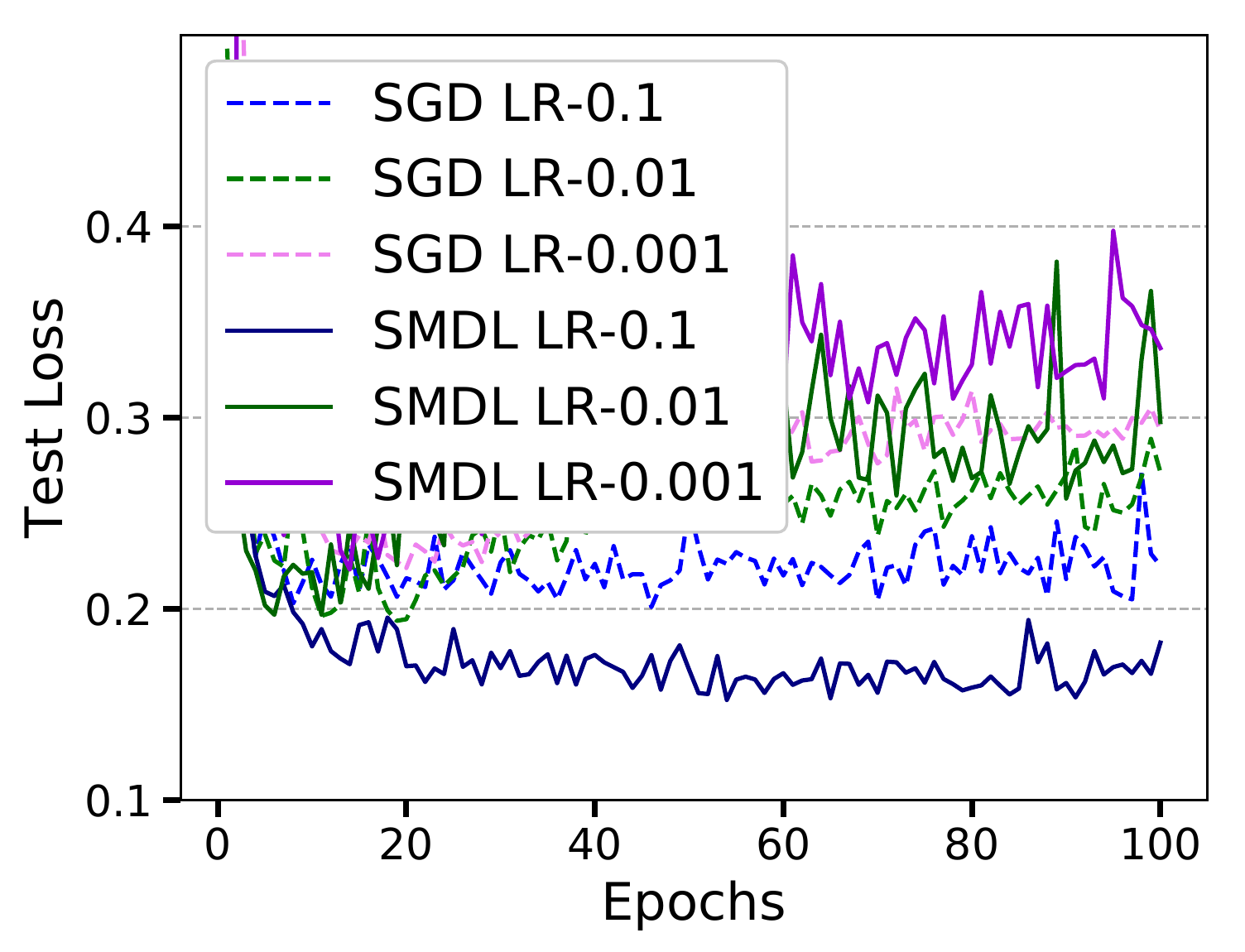}}
\vspace{-10pt}
\subfloat[Test Loss plot with different mini-batch sizes (BS).]{\includegraphics[width=0.49\linewidth]{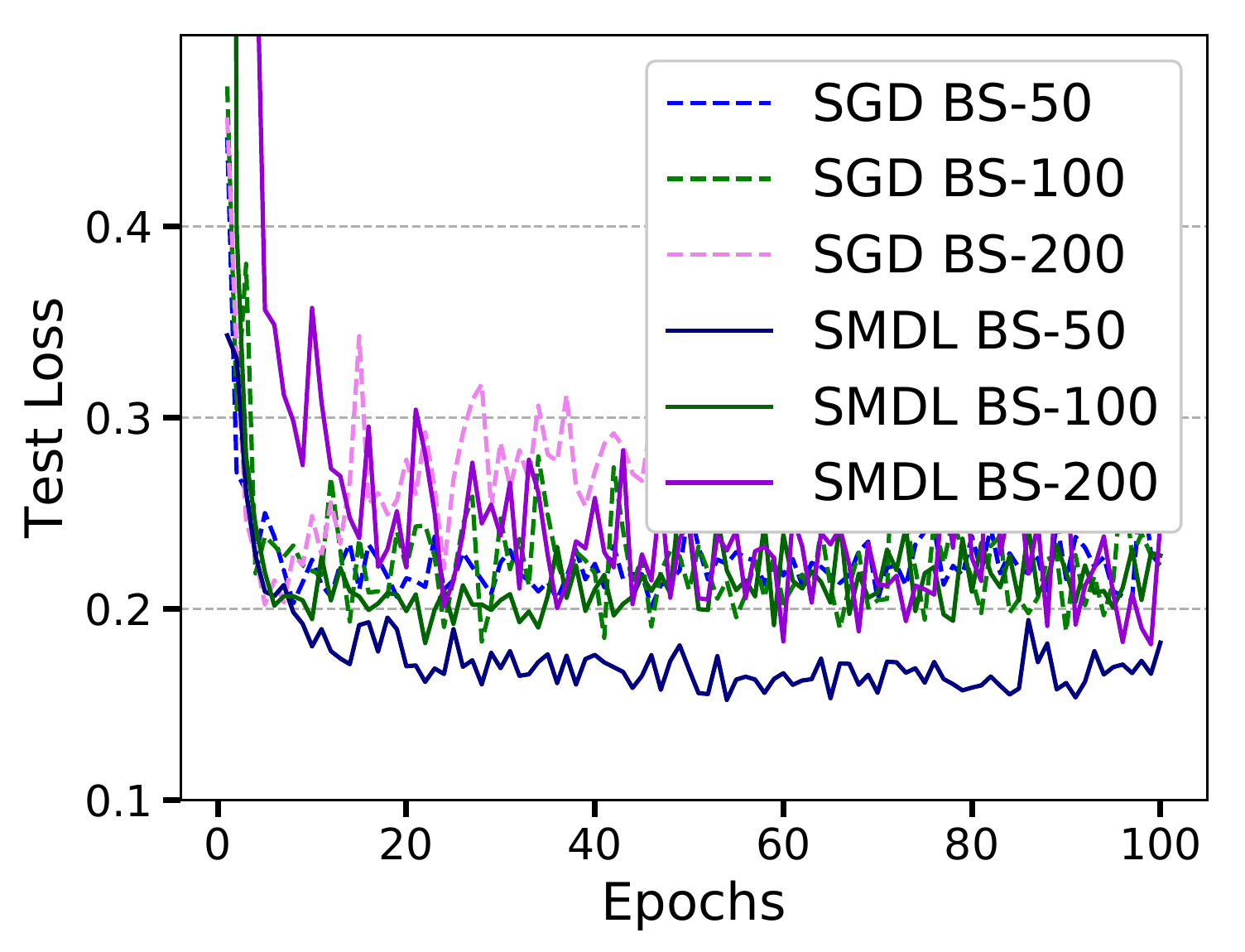}}
\hspace{3pt}\subfloat[Test Loss plot with different Refresh Rates (RR).]{\includegraphics[width=0.49\linewidth]{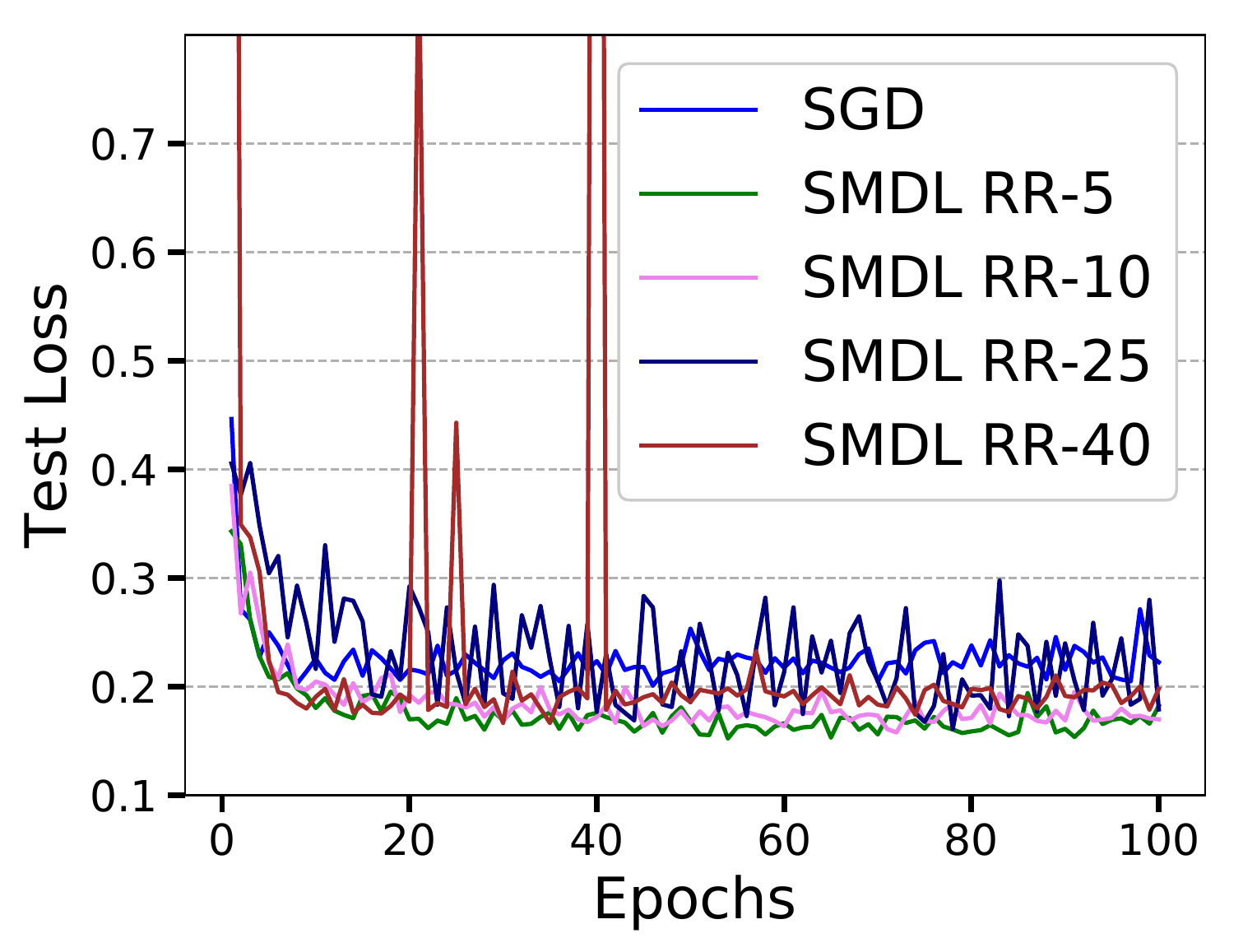}}
\vspace{-2pt}
\caption{The figure shows the ablation results on SVHN dataset.
1)We find that the Euclidean distance measure performs best among other distance metrics. 2)The proposed method SMDL consistently outperforms SGD even with different batch sizes and different learning rates. We note that a batch size of $50$ and a learning rate of $0.1$ gives the least loss for SMDL. 3) We find that a refresh rate of 5 gives the best performance. (refer Section \ref{sec:r_implementation_details}).
}
\label{fig:ablation_99}
\vspace{-6pt}
\end{figure}
%%%%%%%%% FIGURE ENDS

\section{Additional ablation results on trade-off parameters}
\label{sec:additional_trade_off}
The submodular formulation  (Equation \ref{eqn:full}), is a linear combination of four terms, controlled by trade-off parameters. We study the effect of the values for these parameters in Section \ref{sec:r_trade_off_param}. We further do one more ablation to find out whether each of the term is really important for the superior performance of the proposed mini-batch selector.
To study this, we train a ResNet 20 on SVHN dataset four times, with only one of the four terms set to one and others to zero. These models are labeled SMDL-\{1,2,3,4\} in the Figure \ref{fig:ablation_2}. It is compared against the standard SMDL ($\lambda_1 = \text{0.2}, \lambda_2 = \text{0.1}, \lambda_3 = \text{0.5}, \lambda_4 = \text{0.2}$) and SGD.

Figure \ref{fig:ablation_2} shows the result of the experiment. It is evident that the red line, which represents SMDL with contributions from all the terms achieves better generalization performance across epochs.

\section{Test loss plots for various ablation results}

Figure \ref{fig:ablation} in Section \ref{sec:r_lr_batch_size} plots the test error across epochs. Here we plot the test loss for the exact same experiments in Figure \ref{fig:ablation_99}. We note that Euclidean distance metric works best. SMDL consistently outperforms SGD even with different batch sizes and different learning rates. A batch size of 50, learning rate of 0.1 and a refresh rate of 5 gives best performance.   

\section{Training error and loss comparison}
\begin{figure*}
% \vspace{-40pt}
\centering
\subfloat[Train Error on SVHN]{\includegraphics[width=0.30\linewidth]{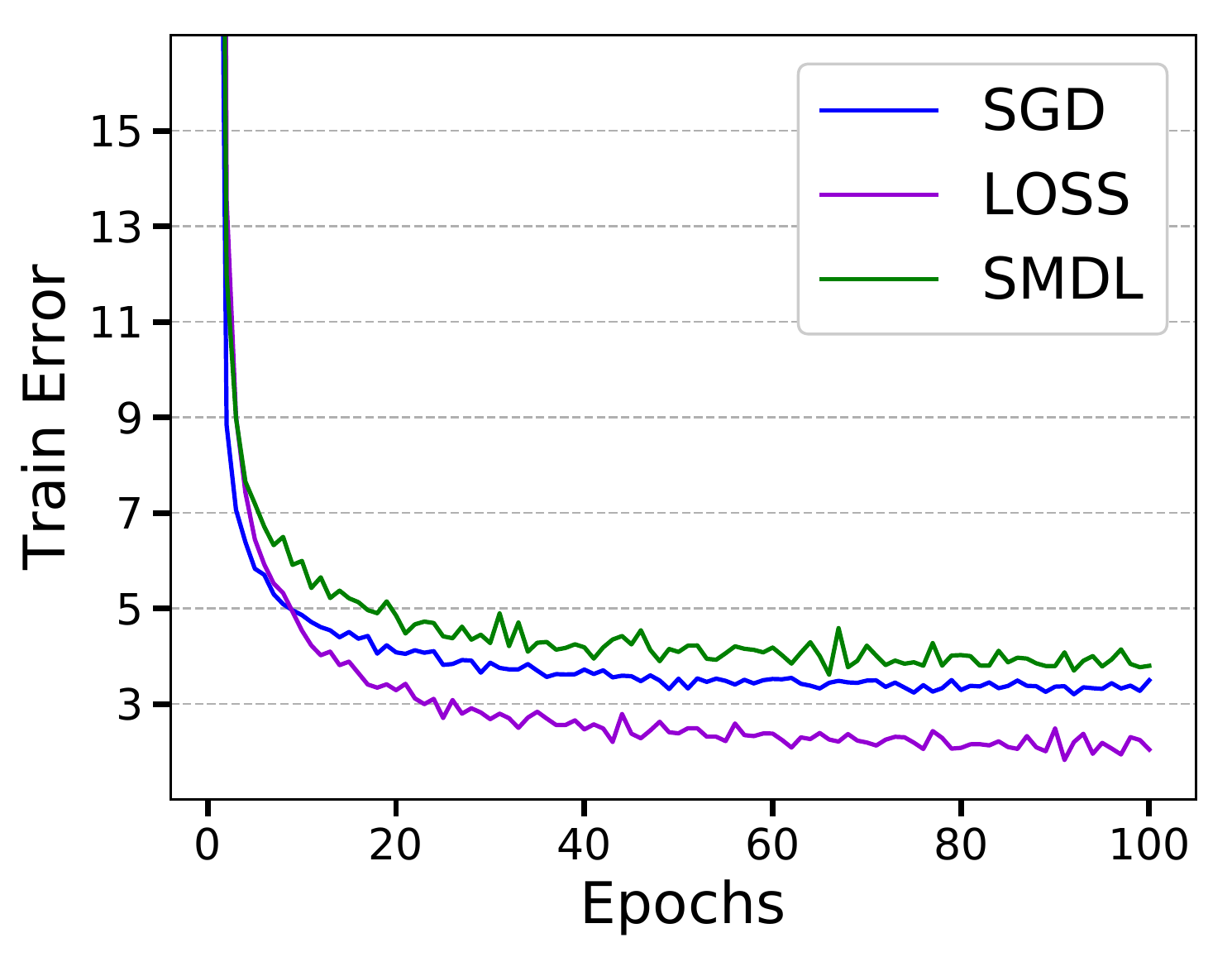}}
\hspace{15pt}\subfloat[Train Error on CIFAR-10]{\includegraphics[width=0.30\linewidth]{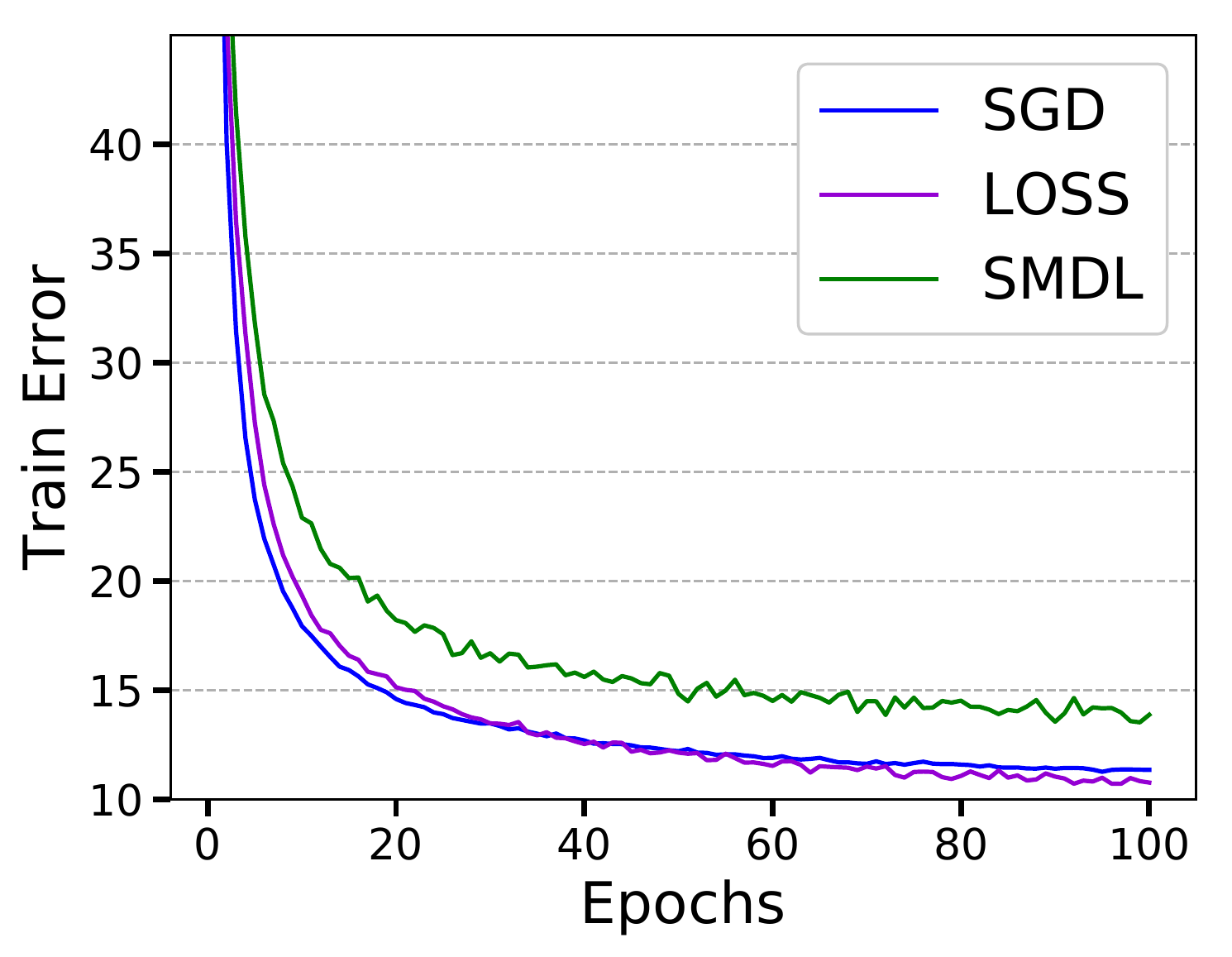}}
\hspace{15pt}\subfloat[Train Error on CIFAR-100]{\includegraphics[width=0.30\linewidth]{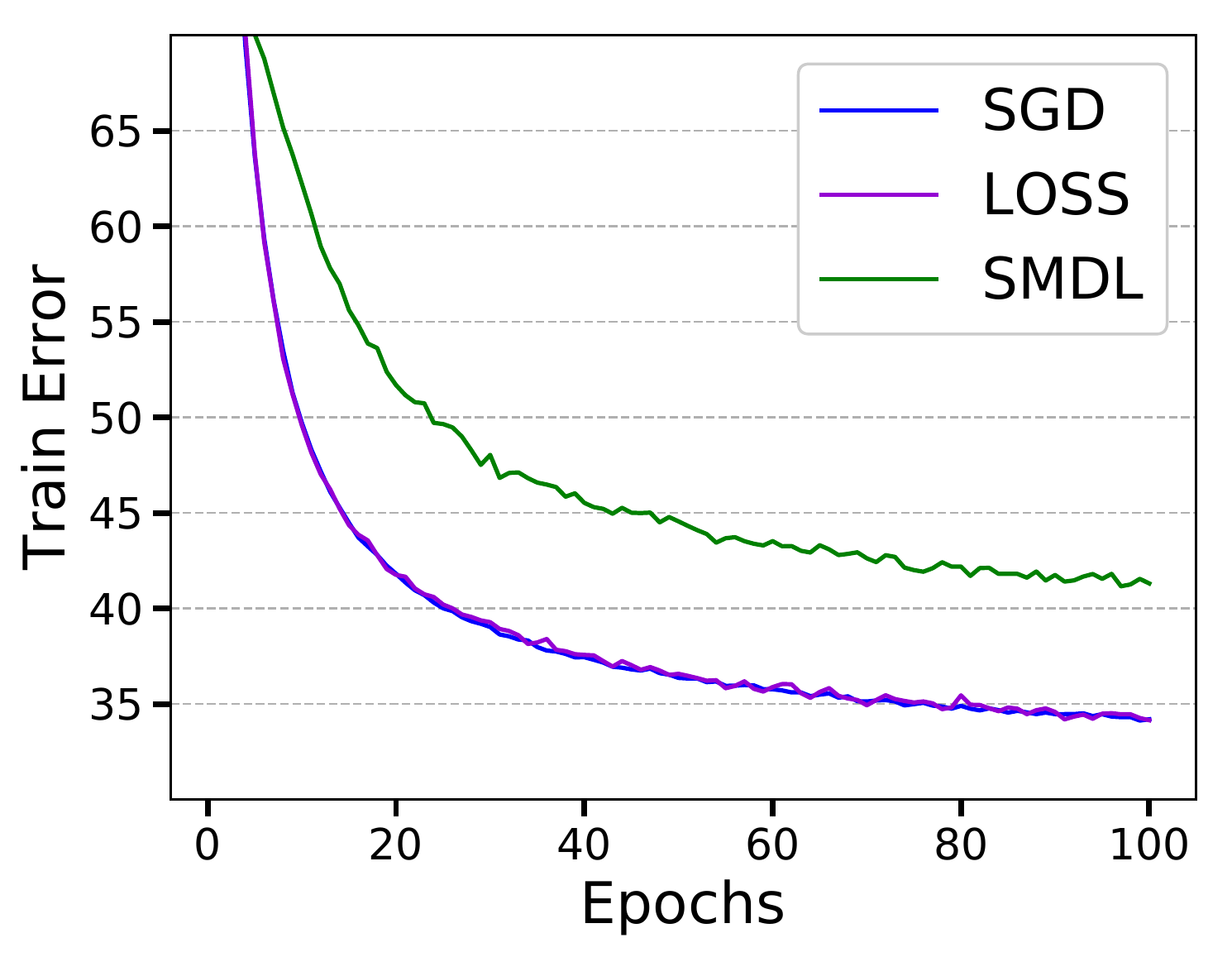}} 
\vspace{-9.6pt}
\subfloat[Train Loss on SVHN]{\includegraphics[width=0.30\linewidth]{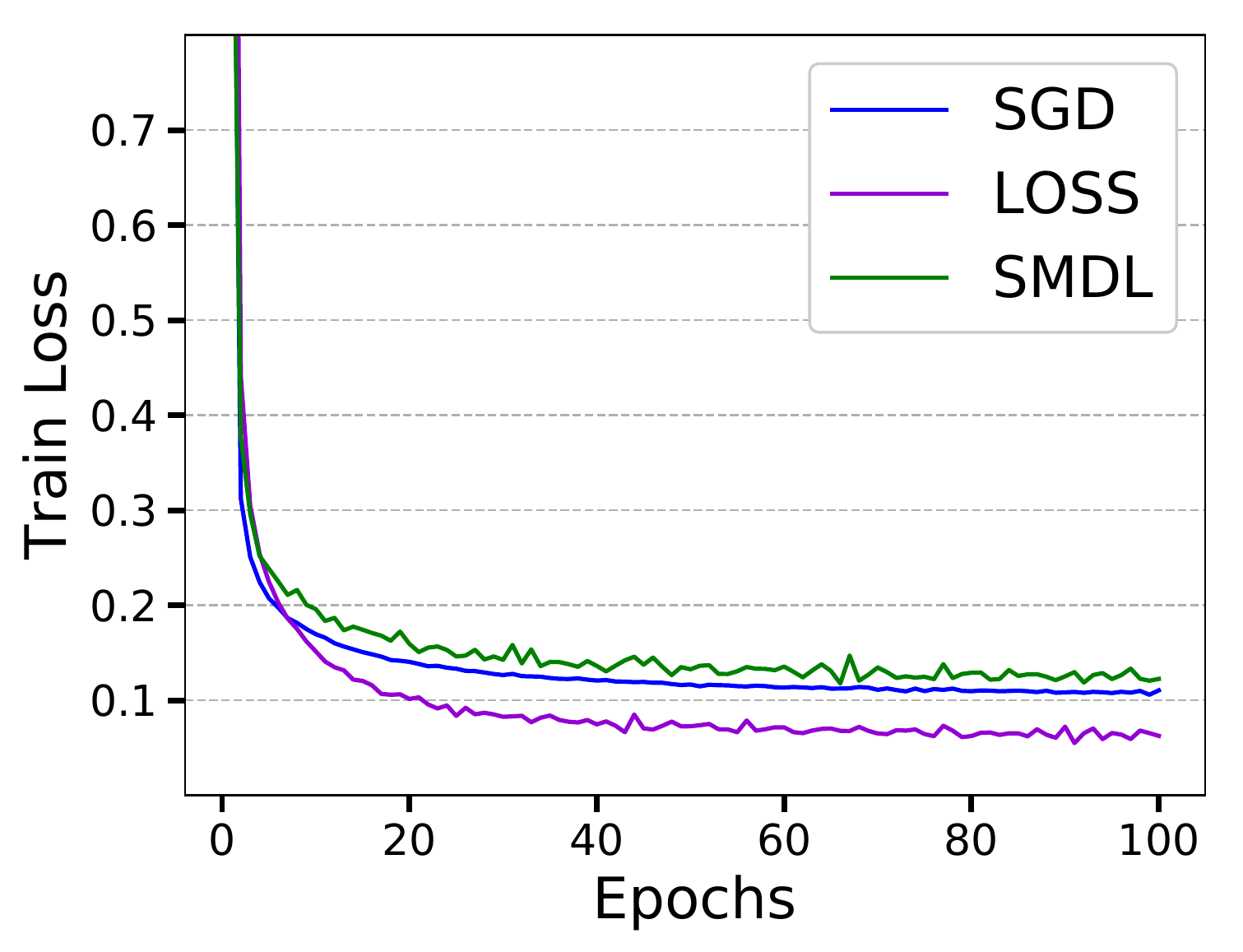}}
\hspace{15pt}\subfloat[Train Loss on CIFAR-10]{\includegraphics[width=0.30\linewidth]{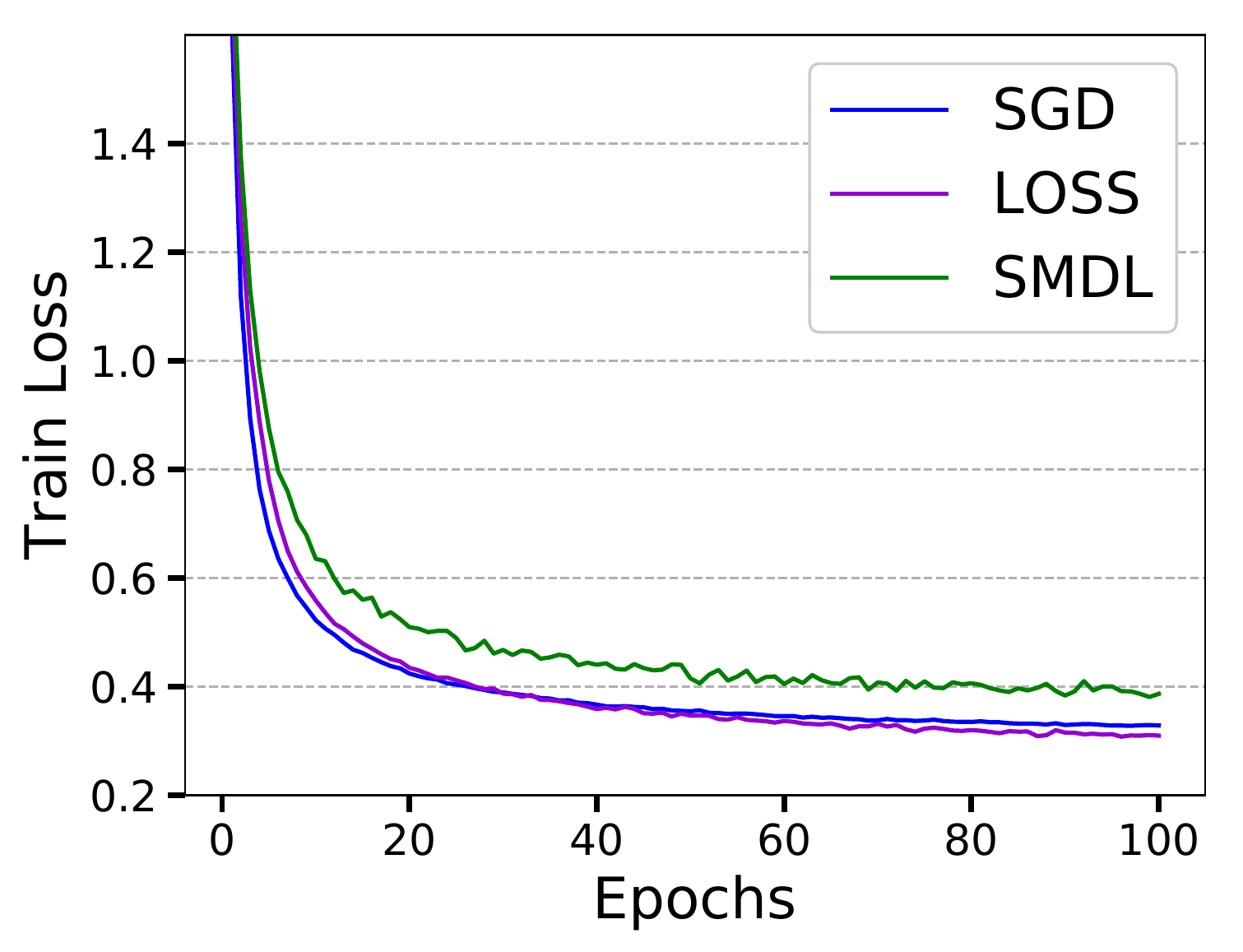}}
\hspace{15pt}\subfloat[Train Loss on CIFAR-100]{\includegraphics[width=0.30\linewidth]{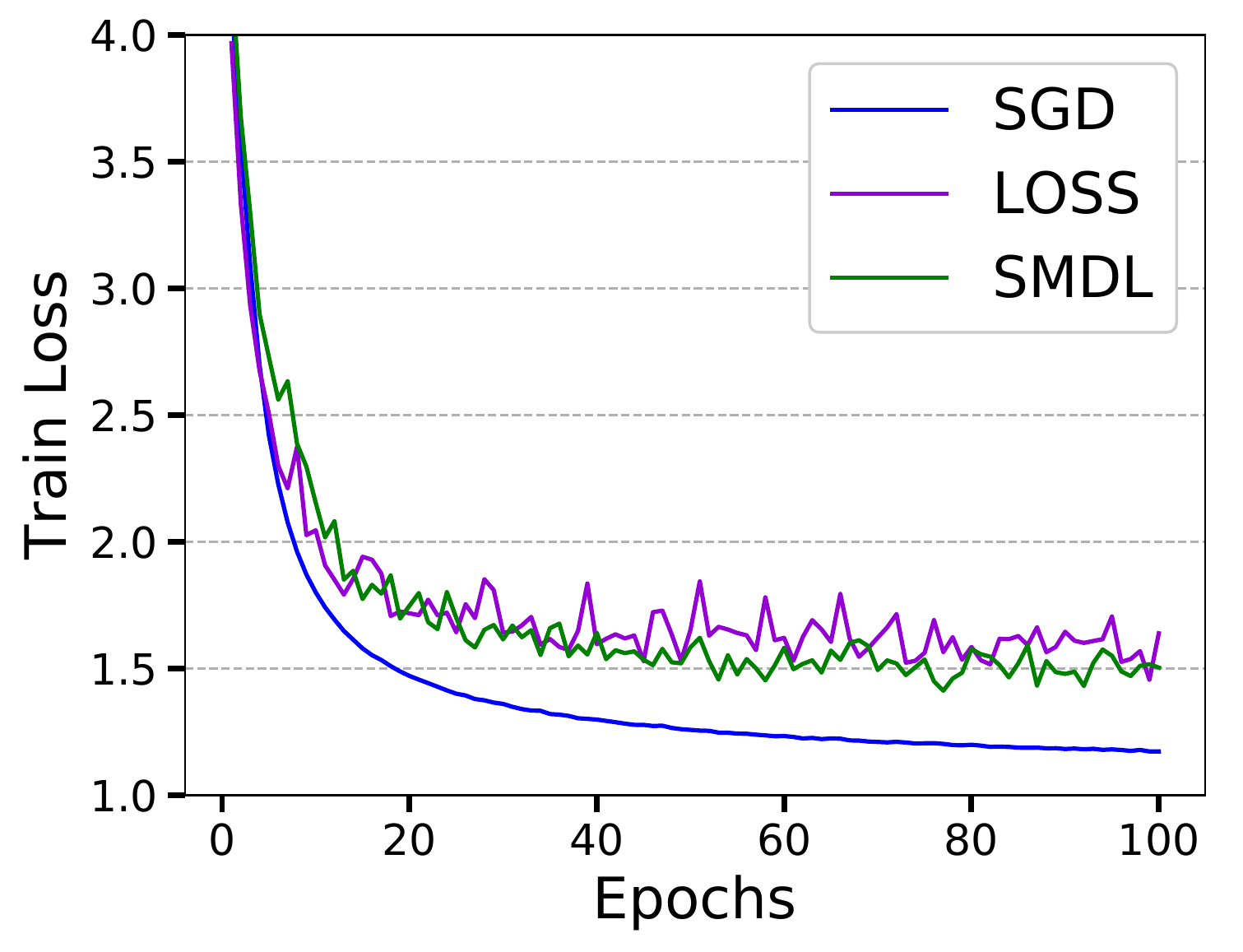}}
% \vspace{-5pt}
\caption{The figure shows how the training error and loss varies over epochs on different datasets. It is very interesting to note that on all the datasets, the training error is much higher for SMDL while the error on the test set is much lower for SMDL (Figure \ref{fig:accuracy_error_plot}). This means that SMDL is not over-fitting and has better generalization capability than SGD and Loss based sampling.}
% \vspace{-10pt}
\label{fig:train_error_loss}
\end{figure*}

We plot the error and loss of the model on the training data across epochs in Figure \ref{fig:train_error_loss}. It is very interesting to see that SGD and Loss based sampling methods have lower training error. 
% This means that the model is over-fitting to the training data.
When we read these graphs along with the results in Figure \ref{fig:accuracy_error_plot}, where error and loss on the test set is plotted, we can see that models trained with SMDL batch selection strategy, has higher error on the training set and lower error on the test set. This indicates that the models trained with SMDL \textit{over-fit less} to the training data and  has \textit{better generalization} capabilities.

% %%%%%%%%% FIGURE STARTS
% \begin{figure}
% \centering
% \vspace{-29pt}
% \subfloat[SVHN: Training error against epochs.]{\includegraphics[width=0.49\linewidth]{plots/SVHN_Train_Error__Main_.pdf}}
% \hspace{3pt}\subfloat[SVHN: Training loss against epochs.]{\includegraphics[width=0.49\linewidth]{plots/SVHN_Train_Loss__Main_.pdf}}
% \vspace{-10pt}
% \subfloat[CIFAR-10: Training error against epochs.]{\includegraphics[width=0.49\linewidth]{plots/CIFAR_10_Train_Error_Main_.pdf}}
% \hspace{3pt}\subfloat[CIFAR-10: Training loss against epochs.]{\includegraphics[width=0.49\linewidth]{plots/CIFAR_10_Train_Loss_Main_.pdf}}
% \vspace{-10pt}
% \subfloat[CIFAR-100: Training error against epochs.]{\includegraphics[width=0.49\linewidth]{plots/CIFAR_100_Train_Accuracy__Main_.pdf}}
% \hspace{3pt}\subfloat[CIFAR-100: Training loss against epochs.]{\includegraphics[width=0.49\linewidth]{plots/CIFAR_100_Train_Loss__Main_.pdf}}
% \vspace{-2pt}
% \caption{The figure shows how the training error and loss varies over epochs on different datasets. It is very interesting to note that on all the datasets, the training error is much higher for SMDL while the error on the test set is much lower for SMDL (Figure \ref{fig:accuracy_error_plot}). This means that SMDL is not over-fitting and has better generalization capability than SGD and Loss based sampling.}
% \label{fig:train_error_loss}
% \vspace{-6pt}
% \end{figure}
% %%%%%%%%% FIGURE ENDS

%%%%%%%%% FIGURE STARTS
%%%%%%%%% FIGURE ENDS

\end{document}